\documentclass[a4paper,reqno]{amsart}  

\usepackage[T1]{fontenc}         
\usepackage[applemac]{inputenc} 
\usepackage{amsthm}
\usepackage{amsmath}
\usepackage{amsfonts}
\usepackage{amssymb}
\usepackage[arrow, matrix, curve]{xy}
\usepackage{picinpar, graphicx}
\usepackage{geometry}
\usepackage{epic}
\usepackage{color}
\usepackage{multicol}
\usepackage{multirow}
\usepackage{wasysym}
\usepackage{url}
\usepackage{float}

\usepackage{lscape}
\usepackage{rotating}

\theoremstyle{remark}

\begin{document}

\renewcommand{\thefootnote}{}

\title[Classification Algorithms and Application to the Framingham Heart Study]
{Comparison of Machine Learning Classification Algorithms and Application to the Framingham  Heart Study}


\author{Nabil Kahouadji}
\date{}

\begin{abstract}  

The use of machine learning algorithms in healthcare can amplify social injustices and health inequities.  While the exacerbation of biases can occur and compound during the problem selection, data collection, and outcome definition, this research pertains to some generalizability impediments that occur during the development and the post-deployment of machine learning classification algorithms. Using the Framingham coronary heart disease data as a case study, we show how to effectively select a probability cutoff to convert a regression model for a dichotomous variable into a classifier. We then compare the sampling distribution of the predictive performance of eight machine learning classification algorithms under four training/testing scenarios to test their generalizability and their potential to perpetuate biases. We show that both the Extreme Gradient Boosting, and Support Vector Machine are flawed when trained on an unbalanced dataset. We introduced and show that the double discriminant scoring of type I is the most generalizable as it consistently outperforms the other classification algorithms regardless of the training/testing scenario. Finally, we introduce a methodology to extract an optimal variable hierarchy for a classification algorithm, and illustrate it on the overall, male and female Framingham coronary heart disease data. \\

\end{abstract} 


\maketitle

\section{Introduction}
As machine learning (ML) and artificial intelligence (AI) are rapidly proliferating in a many aspects of decision-making in society, there is growing concern regarding their ethical use and their potential to perpetuate existing racial biases as highlighted in predictive policing \cite{WK19, O19}, in mortgage lending practices \cite{LF20},  in financial services \cite{K20} and in healthcare \cite{WPS20, FW18, Wal19, Gal19, Gal20}. At the intersection of health, machine learning and fairness, a comprehensive review \cite{Cal21} of the ethical considerations that arise during model development of machine learning in health has been laid out in five stages: problem selection, data collection, outcome definition, algorithm development, and post-deployment considerations, the latter two of which are the main focus of this present research.  For each of these five stages, there are considerations for machine learning to not only mitigate and/or prevent the exacerbation of existing social injustices, but also to attempt to prevent creating new ones. First, interest and available funding influence the selection of  a research problem, and together with the lack of diversity of the scientific workforce, leads to the exacerbation of existing global, racial and gender injustices \cite{V06, Fal20, Hal19}. Second, biases in data collection arise from two processes that result in a loss of data. On one hand, the type of collected data have been shown to suffer, at varying degrees, from challenges and limitations as illustrated for randomized controlled trials \cite{R15, Tal07, SBL15}, electronic health records \cite{FP18, HB09, Kal19},  and administrative health records \cite{Cal14, Lal16}.  On the other hand, historically underserved groups, which includes low- and middle-income nationals \cite{Aal19, Jal16}, transgender and gender-nonconforming individuals \cite{TransSurvey16}, undocumented immigrants \cite{FB11}, and pregnant women \cite{CM19, TBR14},   are often underrepresented, misrepresented, or missing from health data that inform consequential health policy decision. Third stage in the model pipeline is the outcome definition, which may seem to be a straightforward healthcare task, for example defining whether a patient has a disease,  surprisingly can be skewed by the prevalence of such disease and the way it manifest in some patient populations. One of such instances may occur during clinical diagnosis. For example, the outcome label for developing cardiovascular disease could be defined through the occurrence of specific phrases in clinical notes. However, women can manifest symptoms of acute coronary syndrome differently \cite{Cal07} and receive delayed care as a result \cite{Bal17}. In addition, ambiguities occur as a result of diagnosis codes being leveraged for billing purposes rather than for clinical research \cite{KB05}. Another instance where outcome definition can lead to biases and the exacerbation of inequities is the use of  non-reliable proxies to account for, and predict, a health outcome given that socioeconomic factors affect access to both healthcare and financial resources. The fourth stage in the model pipeline is the algorithm development per se.  Even when all considerations and precautions have been taken in the previous three stages to minimize the infiltration of biases, noise, and errors in the data, the choice of the algorithm is not neutral, and often is the source of impediments to an ethical deployment of the algorithm. The crucial factors in model development are: understanding confounding, feature selection, tuning parameters, performance metrics, and group fairness definition. Indeed, confounding features are those features that influence both the independent and dependent variables, and as the vast majority of models learn patterns based on observed correlations between the training dataset, even when such correlations do not occur in the testing dataset, it is critical to account for confounding features as illustrated in the classification models designed to detect hair color \cite{Jal18},  and in predicting pneumonia risk and hospital 30-day readmission \cite{Cal15}. Moreover,  blindly  incorporating  factors like race and ethnicity, which are increasingly available due to the large-scale digitization of electronic health records, may exacerbate inequities for a wide range of diagnosis and treatments \cite{Val20}. Therefore, it is crucial to carefully select model's features, and to consider the human-in-the-loop framework where the incorporation of automated procedures is blended with investigator knowledge and expertise \cite{Kal20}.  Another crucial component of the algorithm development is the tuning of the parameters, which can be set a priori, selected via cross-validation, or extracted from a default setting from a software,  are ways that can lead to overfitting of the model to the training dataset, and a loss of generalizability to the target population, the latter of which is a central concern for ethical machine learning.  To assess and evaluate a model, many performance metrics are commonly used, such that area under the receiver operating characteristic curve (AUC),  and area under the precision-recall curve (AUPRC) for regression models on one hand, and accuracy, true positive rates, and precision for classification models on the other hand. It is important to use a performance metric that reflects the intended use case and to be aware of potential misleading conclusions when using so-called objective metrics and scores \cite{Val20}.  Finally, the fifth stage of the model pipeline is the post-deployment of the model in a clinical, epidemiological, or policy service. A robust deployment requires a careful performance reporting, auditing generalizability, documentation, and regulation.  Indeed, it is important to measure and address the downstream impact of models through auditing for bias and examination of clinical impact \cite{CSG19}. It is also crucial to evaluate and audit the deployment of the generalization, as any shift in the data distribution can significantly impact model performance when the setting for development and deployment differ as illustrated in chest X-ray models \cite{Zal18, Lal20, Sal20}.  While some algorithms have been proposed to account for distribution shifts post-deployment \cite{SS20},  their implementation suffer from significant limitations due to the requirement of the specification of the nature or amount of distributional shift, thus rendering a tedious periodic monitoring and auditing.  The last two ingredients for an ethical post-deployment of machine learning are the establishment of clear,  and insightful model and data documentation as well as the adherence to best-practice, and compliance with regulation.

With both the algorithm development, and post-deployment considerations of the five-stage model pipeline\cite{Cal21} in mind, this research focuses on classification algorithms using electronic health records, with the goal of identifying a methodology that is not only ethical, robust,  and understandable by practitioners or community members, but also insensitive to distributional shifts,  and thus more generalizable. While regression models for dichotomous observations are widely used by modelers, the probability outcome for patient-level data may not be insightful and helpful to determine whether a given patient should undergo further intervention, that is, whenever the response of a practitioner to a patient (or family member) is expected to be a binary response rather than a vague probability statement. Regression algorithms can be converted to classification algorithm by choosing a probability cutoff. However, the naive choice of a 50\% probability cutoff often leads to higher misclassifications. Using the Framingham coronary heart disease data as a case study, we first show  how a probability cutoff can be determined to effectively convert  logistic \cite{RAbook} and random forest \cite{RF} regression models into classifiers. We then compare the performance of eight classification algorithms, two of which are widely used supervised machine learning algorithms, namely, Extreme Gradient Boosting (XGB) \cite{XGB}, and Support Vector Machine (SVM)\cite{SVMbook}, together with the logistic and random forest classifiers, and two uncommonly used supervised machine functions, a.k.a, linear and quadratic discriminant functions \cite{DAbook}, and we introduce two different combinations of the linear and quadratic discriminant functions into two scoring functions, namely the double discriminant scoring of types I and II. Using a paired design set up, we perform a sampling distribution analysis for these eight classification algorithms under four different training/testing scenarios, and for varying training/testing ratios. We determine from the  comparison of the performance sampling distributions the algorithm that consistently outperforms and is the least sensitive to distributional shifts. We then layout and illustrate a methodology to extract an optimal variable hierarchy, that is a sequence of variables that provides the most robust and most generalizable variable selection for the classification algorithm given a performance metric. For instance, if the optimal variable hierarchy for a classification algorithm and a performance metric is a sequence of two variables, then the first variable is the optimal single variable among the set of all features, the first and second variables constitute the optimal pair of variables among all pairs of features, and the inclusion of any extra feature(s) to this 2-variable hierarchy would decrease the performance metric. We show in particular that this optimal variable hierarchy satisfies the Bellman principle of optimality, leading then to the reduction of sampling distribution tests from $2^p-1$ iterations to at most $p(p+1)/2$, where $p$ is the number of variables (features). This methodology is applied to entire Framingham CHD data, and to both male and female CHD data.




\section{Materials and Methods}




The Framingham Heart Study is a widely acknowledged premier longitudinal cohort study \cite{Fram1, Fram2, Fram3, FramLast}. Motivated by the mounting epidemic of cardiovascular disease (CVD) in the 1950s, and becoming the leading cause of death, and the reason why life expectancy beyond age 45 did not increase, action was needed to identify the determinants of the disease process, and given that no treatment capable of prolonging life for those who managed to survive an attack existed, a preventive approach was deemed more important than a search for a cure.  Moreover, given that CVD is a disease that develop over time, a longitudinal study was necessary.  The Framingham Study \cite{Fram1, FramLast} was conducted as follows: a systematic sample of 2 of every 3 families in the town of Framingham, Massachusetts, was selected. People in those families between the ages of 30 and 59 years were invited to participate in the study. A total of 5,209 individuals (2,336 men and 2,873 women) joined the study with the goal of collecting epidemiological data on CVD, and the establishment of the relation of risk factors such as clinical  (age, sex, blood pressure, cholesterol, body weight and diabetes), and lifestyle  (smoking, physical activity, and alcohol consumption) parameters. Participants of the Framingham Study were continuously monitored to identify when a CVD event occurred. Given the success of the Framingham Study, a second cohort in 1971 with over 5,000 subjects, and a third cohort in 2001 with over 4,000 subjects, led to two replications \cite{Fram4, Fram5}. The reader may refer to \cite{FramLast} for the most recent review of the Framingham Study and an overview of the Framingham risk functions for CVD and coronary heart disease (CHD). Note that CHD include myocardial infarction (a.k.a., heart attack), coronary death, stroke and heart failure.

Using the Framingham coronary heart disease data available on Kaggle, we extracted a sub-data consisting of seven explanatory variables representing for each patient the age $X_1$, total cholesterol $X_2$, systolic blood pressure $X_3$, diastolic blood pressure $X_4$, body mass index (BMI) $X_5$, heart rate $X_6$ and number of cigarettes smoked per day $X_7$, and  one dichotomous response variable $Y$ representing whether a patient had a coronary heart disease in the 10-year period following the measurement of these seven explanatory variables. The sub-data extracted, which we refer to as the Framingham CHD data in this paper, consists of all patients for which there is no missing measurement for each of the seven explanatory variables. The goal of this study is to compare the sampling distribution of the performance of several machine learning classification algorithms under several training/testing scenarios to  1) determine a methodology that will better predict whether (or not) a patient will develop a coronary heart disease in the 10-year following the measurement of the seven explanatory variables, 2) determine the best way to train these machine learning algorithms,  and their sensitivity to both the size and the distribution of the training datasets, and 3) extract an optimal variable hierarchy for the explanatory variables, and assess their generalizability to new medical and geographical data.

 This Framingham CHD data consist then of two groups: Group 1 with $N_1 = 622$ patients who had a coronary heart disease in the 10-year period following the beginning of the study, and Group 2 with $N_2 = 3520$ patients who did not have a coronary heart disease in the same 10-year period. The prevalence of CHD in this data is then $15\%$, meaning that 15\% of the $N = N_1 + N_2 = 4142$ patients  had a coronary heart disease. Data analysts often split randomly the data into a training and testing datasets using a training ratio $\tau$. For instance, if one assigns randomly 80\% of the data to the training dataset, and then assign the remaining 20\% of the data to the testing dataset, then the training ratio is $\tau = 0.8$. Given a training ratio $\tau$, let's say $\tau = 0.8$, there are several ways to split the data into training and testing sets, the simplest of which is to merge Group 1 and Group 2, and then randomly split the data into training and testing datasets using a given training ratio $\tau$. This simple splitting leads to an over representation in the training dataset of the largest of the two groups, for instance Group 2 in this Framingham CHD data. For reasons that will be clarified in the findings of subsection \ref{comparealgo}, we do not consider this simple splitting in this paper, but rather the following four training/testing scenarios. Let's denote by $n_1$ the number of observations in the intersection of Group 1 and the training dataset, by $n_2$ the number of observations in the intersection of Group 2 and the training dataset, and by $n_3$ the number of observations used in the testing dataset. Therefore, the number of observations in the training dataset is $n_1+n_2$. In what follows, $[.]$ denotes the rounding to the nearest natural number. For a fixed training ratio $\tau $, let's consider these four training/testing scenarios: 
  \begin{enumerate}
\item Proportional training and testing: randomly select $n_1 = [ \tau N_1]$ observations from Group 1, and  $n_2 = [\tau N_2]$ observations from Group 2, to form the training dataset, and then use the remaining $n_3 = [(1-\tau) (N_1+N_2)]$ observations for the testing dataset.  
\item Equal training and proportional testing: randomly select $n_1 = n_2 = [ \tau  \min(N_1, N_2)]$ observations from each Groups 1 and 2 to form the training dataset, then  randomly select $(1-\tau) N_1$ from the remaining observations of Group 1, and $(1-\tau) N_2$ from the remaining observations of Group 2, to form the testing dataset of size $n_3 = [(1-\tau) (N_1+N_2)]$. 
\item Proportional training and equal testing: randomly select $n_1 = [ \tau N_1]$ observations from Group 1, and  $n_2 = [\tau N_2]$ observations from Group 2 to form the training dataset, then randomly select $(1-\tau)\min(N_1, N_2)$ from each of Groups 1 and 2 to form the testing dataset, and hence $n_3 = 2(1-\tau)\min(N_1, N_2)$. 
\item Equal training and testing: randomly select $n_1 = n_2 = [ \tau  \min(N_1, N_2)]$  from each Groups 1 and 2 to form the training dataset, and  then to form the testing dataset, randomly select  $ [(1-\tau)\min(N_1, N_2)]$ observations from the remaining observations for each of Groups 1 and 2, and hence  $n_3 = 2[(1-\tau)\min(N_1, N_2)]$. 
\end{enumerate}
Using the Framingham CHD data, where $N_1 = 622$, $N_2 = 3520$ and with a training ratio $\tau = 0.8$, the sizes $n_1$, $n_2$ and $n_3$ for each of the four training/testing scenarios are given in Table \ref{Sizen1n2n3}.

\begin{table}[htp]
\begin{center}
\begin{tabular}{llrrr}
\hline
Training & Testing & $n_1$ & $n_2$ & $n_3$\\ \hline
Proportional & Proportional & 498 & 2819 & 124 + 704 = 828\\
Equal & Proportional & 498 & 498 & 124 + 704 = 828\\
Proportional & Equal & 498 & 2819 & 124 + 124 = 248\\
Equal & Equal & 498 & 498 & 124 + 124 = 248\\
\hline
\end{tabular}
\end{center}
\caption{Training and testing datasets sizes across four training/testing scenarios  and for a training ratio $\tau = 0.8$}
\label{Sizen1n2n3}
\end{table}%


Once a classification model has been tested, one can produce the confusion matrix, that is, a table specifying the frequencies of the true positive (TP), the false positive (FP), the false negative (FN) and the true negative (TN) predictions, and then use this confusion matrix to compute one or several performance metrics to assess the classification model such as the accuracy, true positive rate (sensitivity), true negative rate (specificity) and precision.  In what follows, we recall the definitions (and formulas) of the accuracy, true positive and true negative rates. We then  introduce and make a distinction between positive precision and negative precision on one hand, and  between the observed prevalence and the expected prevalence on the other hand. Finally, we introduce  the \textit{Classification Performance Matrix} as an extension of the standard confusion matrix along with the above seven model performance metrics. Let's refer to the total number of predictions as the \textit{Grand Total}, i.e., $\text{Grand Total = TP + FP + FN + TN}$.  Recall that the \textit{Accuracy} (Acc) of a classification model is the ratio of the correct predictions among the total number of predictions, i.e., $Acc = (TP+TN)/\text{Grand Total}$. Recall also that the \textit{True Positive Rate} (TPR), also called \textit{sensitivity} in the literature, is the ratio of the number of true positive predictions among the total number of actual positive tested cases, i.e., $TPR = TP / (TP + FN)$. Finally, recall that the \textit{True Negative Rate} (TNR), also called \textit{specificity} in the literature, is the ratio of the number of true negative predictions among the total number of actual negative tested cases, i.e., $TNR = TN / (FP + TN)$. We define the \textit{Positive Precision} (PPrec), which is simply called \textit{precision} in the literature, as the ratio of the number of true positive predictions among the total number of tested cases that have been predicted as positive by the classification model, i.e., $PPrec = TP / (TP + FP)$. Similarly,  we define the \textit{Negative Precision} (NPrec) as the ratio of the number of true negative predictions among the total number of tested cases that have been predicted as negative by the classification model, i.e., $NPrec = TN / (FN + TN)$. We define the \textit{Observed Prevalence} (OPrev), which is simply called \textit{prevalence} in the literature, as the proportion of positive cases among the tested cases, i.e., $OPrev = (TP+FN)/\text{Grand Total}$. Finally, we defined the \textit{Expected Prevalence} (EPrev) as the ratio between the positive predictions among the total number of tested cases that have been predicted as positive by the classification model, i.e., $EPrev = (TP+FP)/\text{Grand Total}$.  Lastly, we introduce an extension of the confusion matrix with the above seven performance metrics to provide a practical and comprehensive way to compare the performance of several classification algorithms across the four training/testing scenarios. This classification performance matrix (Table \ref{classperfmatrix}) allows, at a glance, to detect whether a classification model is biased and flawed. For instance, if the prevalence of the testing data set is high, let's say 95\%, then a classification model that predicts all tested cases as positive would have a 95\% accuracy, and a 100\% true positive rate. A modeler that uses only these two metrics to build and/or compare models would have a false impression of having a great model.  However, this model would be flawed, as the true negative rate is 0\%, which means that all tested cases that are negative would be predicted as positive, and thus it would trigger in a medical context additional interventions that are financially, timely and emotionally costly, rendering the application of this model useless. 

\begin{table}[htp]
\begin{center}
\begin{tabular}{cc|cc|cc}
\cline{3-4}
&&\multicolumn{2}{c|}{Predicted}& \\ \cline{3-6}
&& Positive &Negative & \multicolumn{1}{c|}{Total} & \multicolumn{1}{c|}{True Rate \%} \\ \hline
\multicolumn{1}{|c}{\multirow{2}{*}{Actual} }& \multicolumn{1}{|c|}{Positive} & TP  & FN& \multicolumn{1}{c|}{TP + FN} & \multicolumn{1}{c|}{TPR} \\
\multicolumn{1}{|c}{}&\multicolumn{1}{|c|}{Negative} & FP& TN& \multicolumn{1}{c|}{FP + TN} & \multicolumn{1}{c|}{TNR} \\ \hline
&\multicolumn{1}{|c|}{Total} & TP + FP & FN + TN & \multicolumn{1}{c|}{Grand Total} & \multicolumn{1}{c|}{OPrev} \\ \cline{2-6}
& \multicolumn{1}{|c|}{Precision \%} & PPrec & NPrec & EPrev & \multicolumn{1}{|c|}{Acc}\\ \cline{2-6}
\end{tabular}
\end{center}
\caption{Classification Performance Matrix}
\label{classperfmatrix}
\end{table}%

\section{Results}


\subsection{Logistic and Random Forest Classifiers}\label{LandRF}
\begin{figure}[htbp]
\begin{center}
\includegraphics[scale=.22]{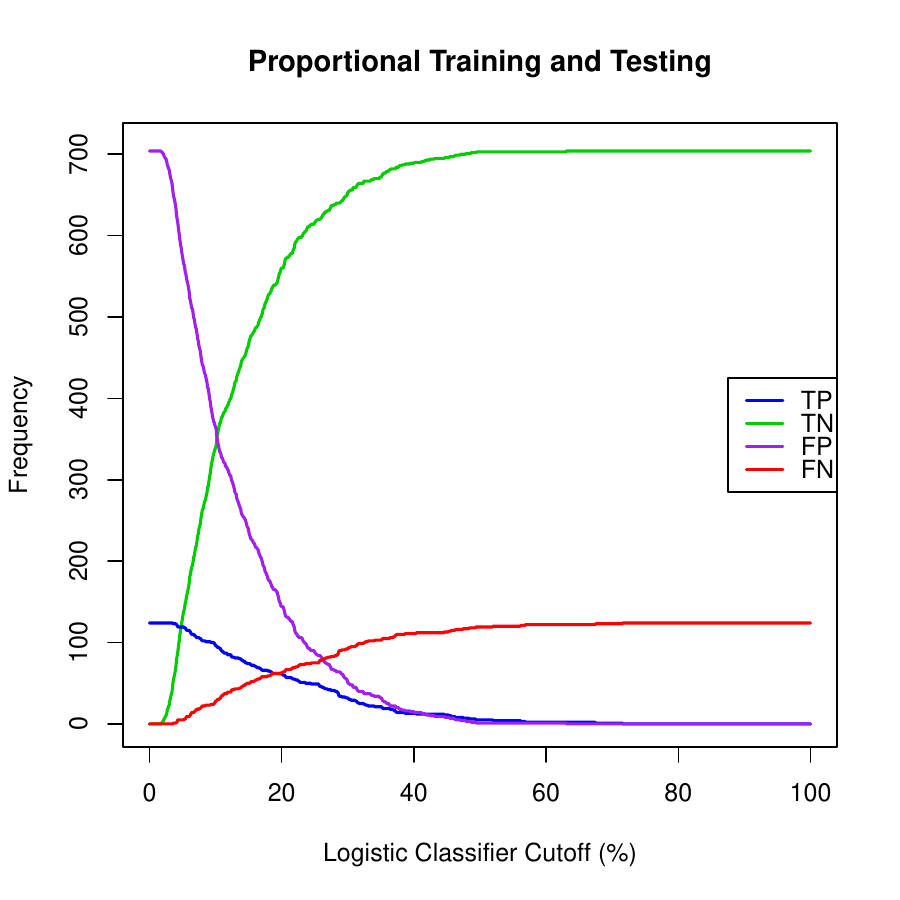}\includegraphics[scale=.22]{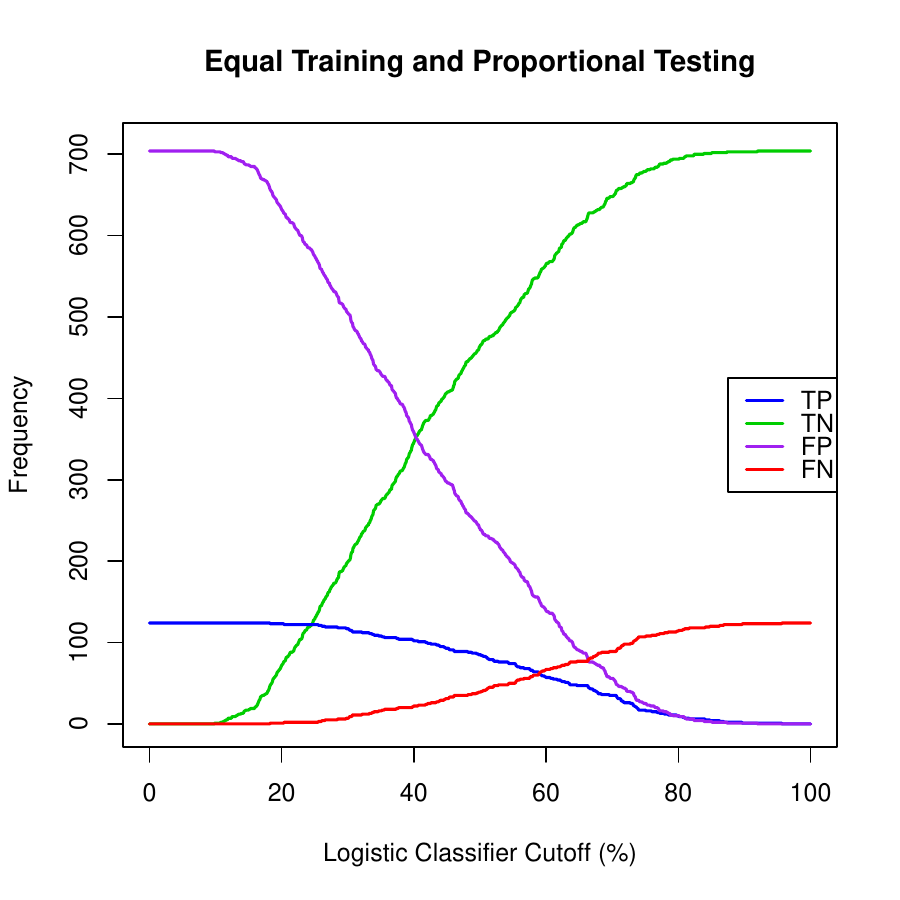}\includegraphics[scale=.22]{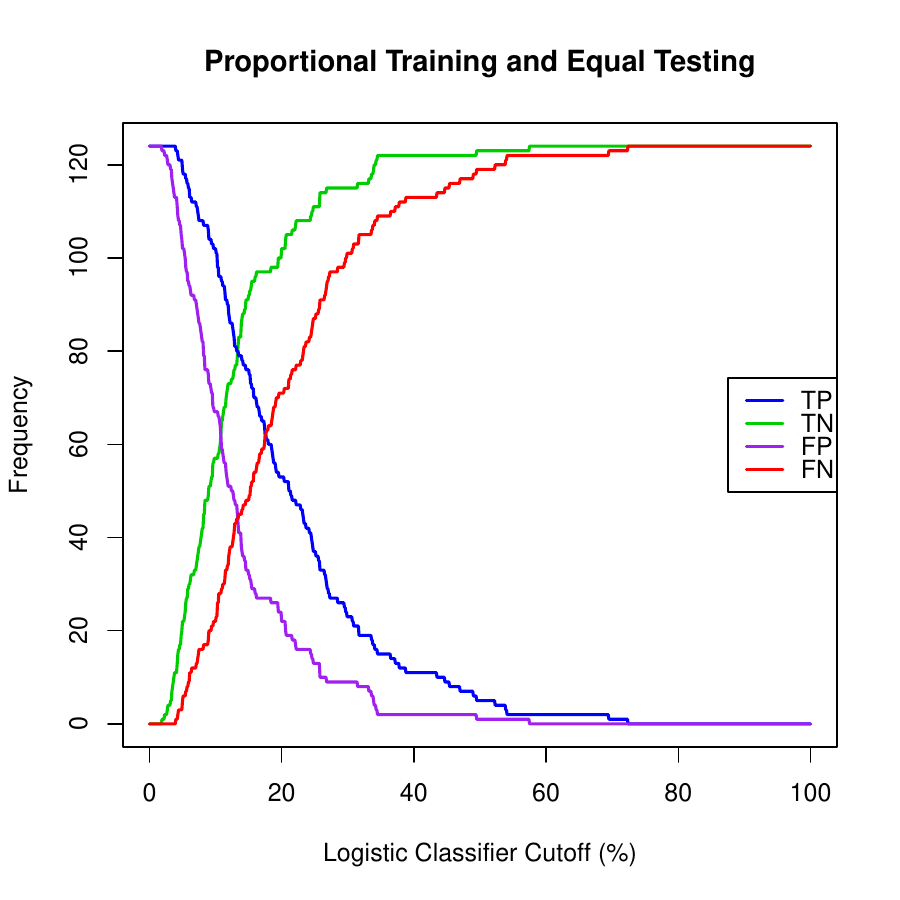}\includegraphics[scale=.22]{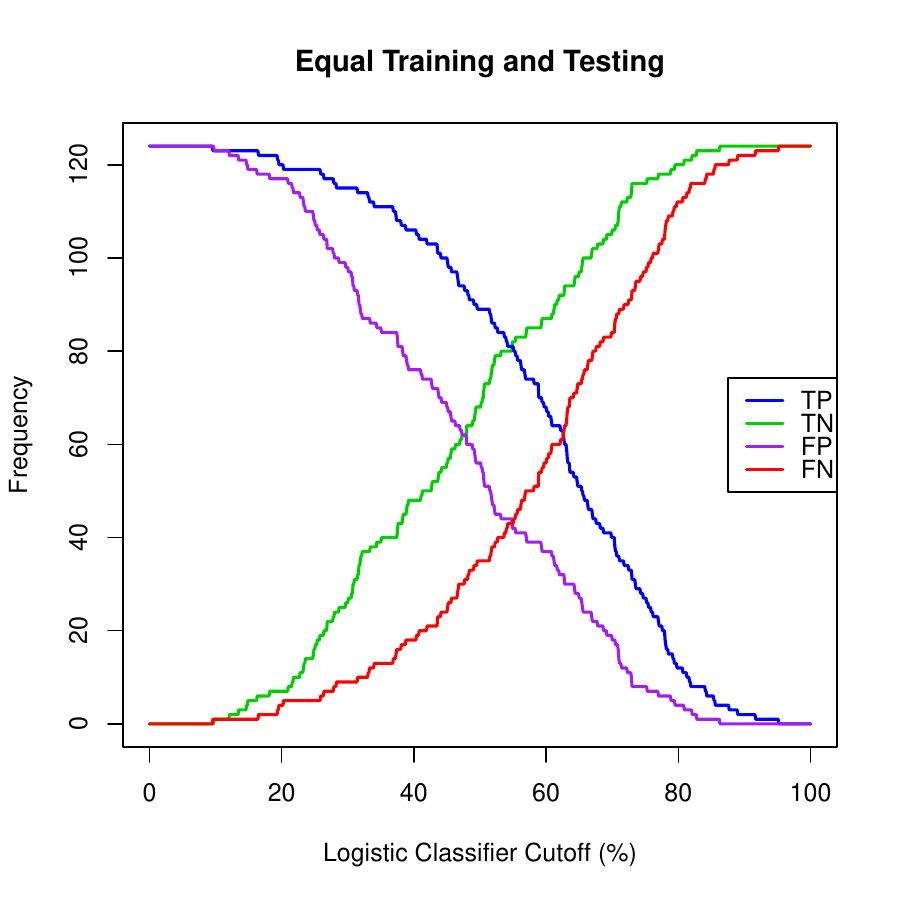}
\includegraphics[scale=.22]{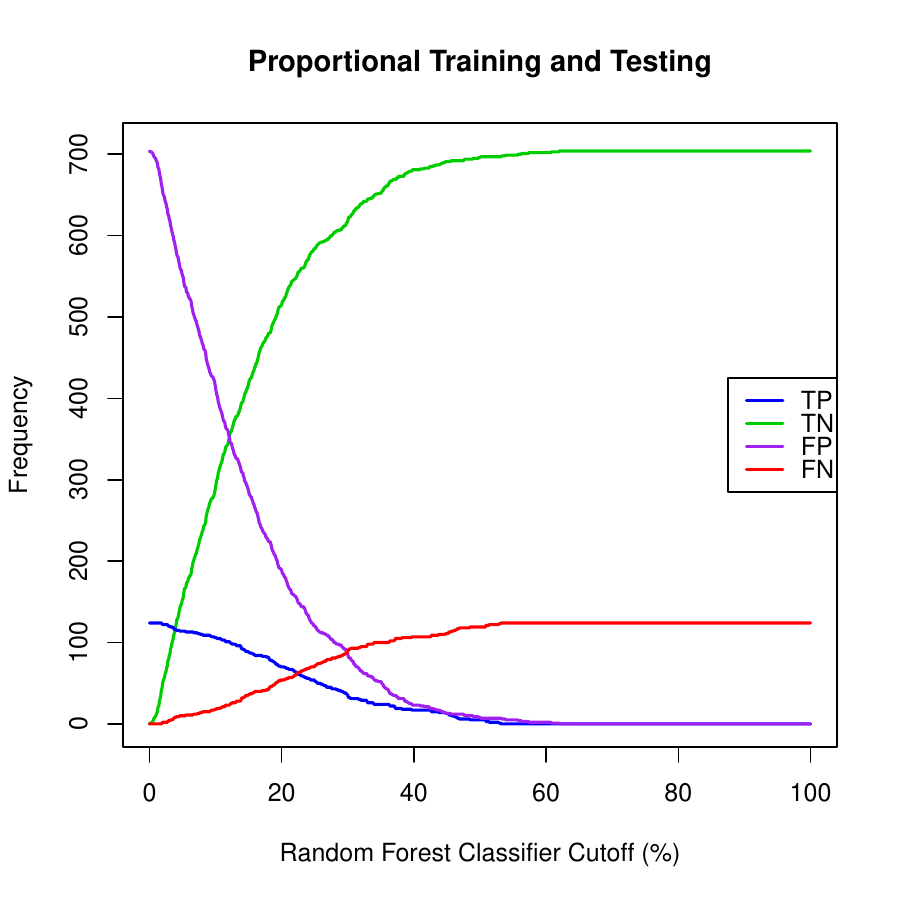}\includegraphics[scale=.22]{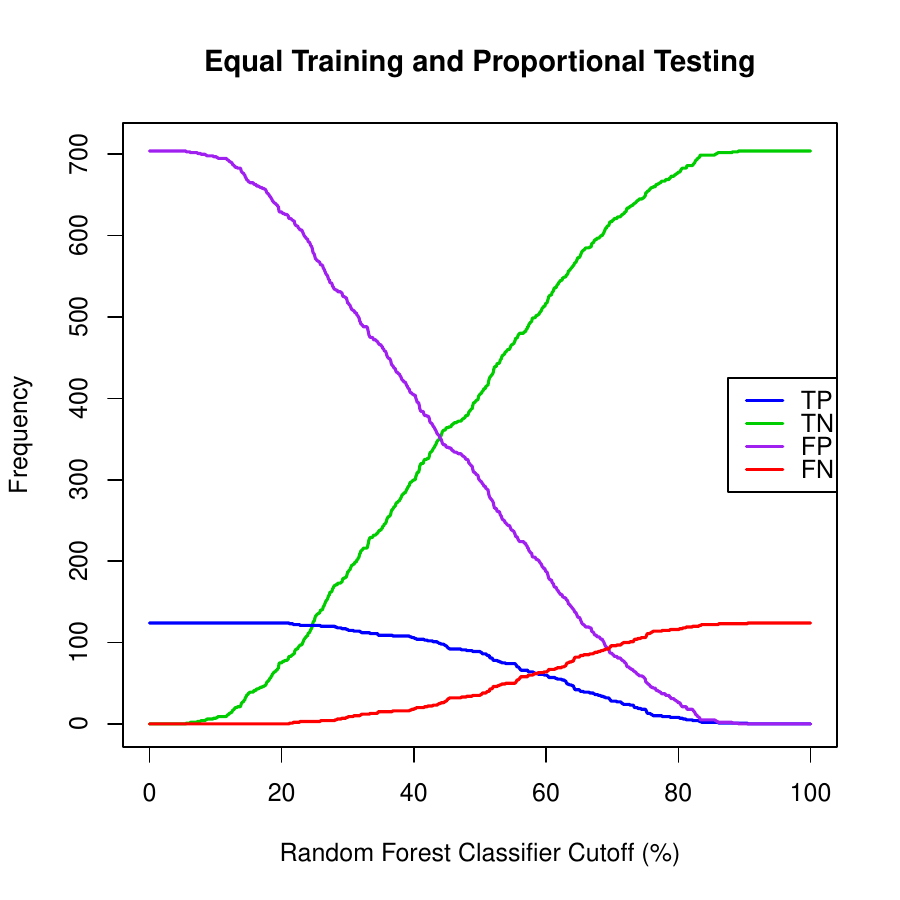}\includegraphics[scale=.22]{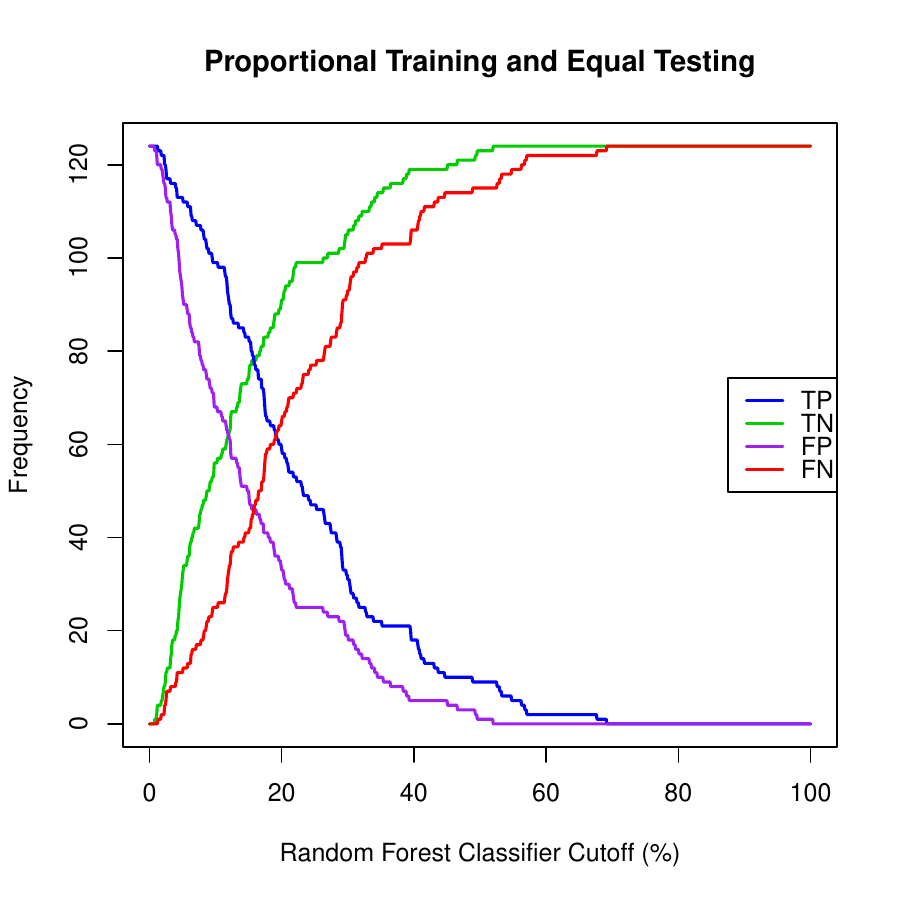}\includegraphics[scale=.22]{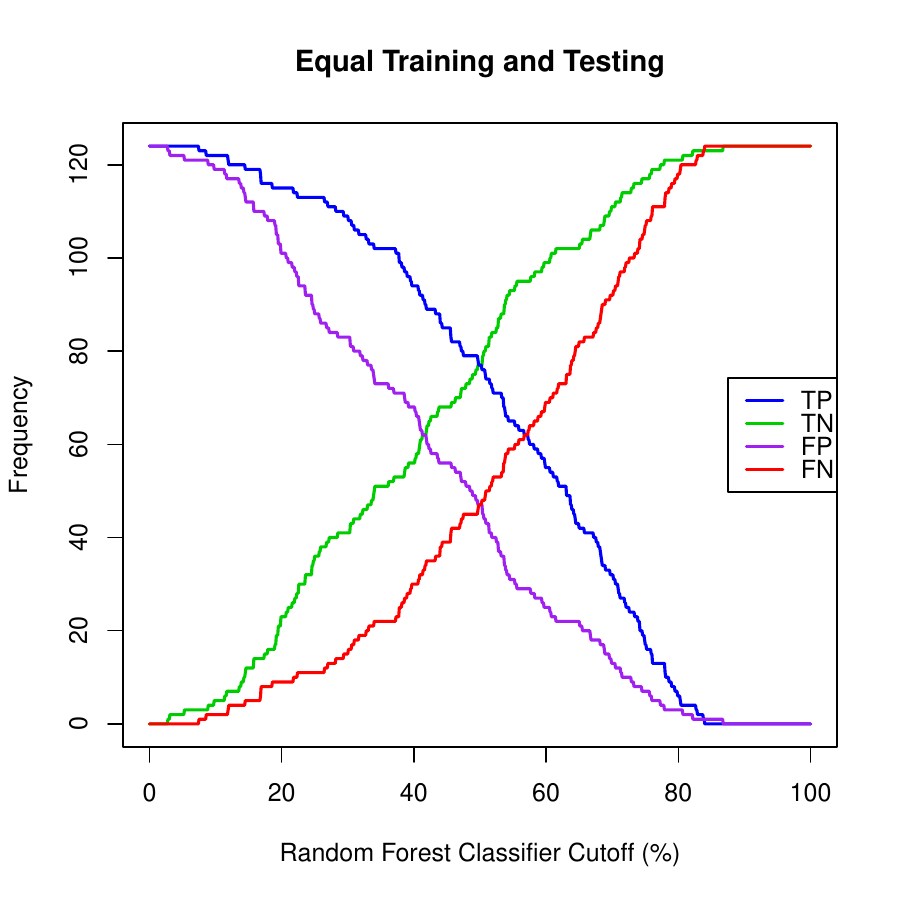}
\caption{One simulation for the Logistic and Random Forest Classifier Cutoffs for four training/testing scenarios}
\label{glmrfOneSim}
\end{center}
\end{figure}

\begin{figure}[htbp]
\begin{center}
\includegraphics[scale=.22]{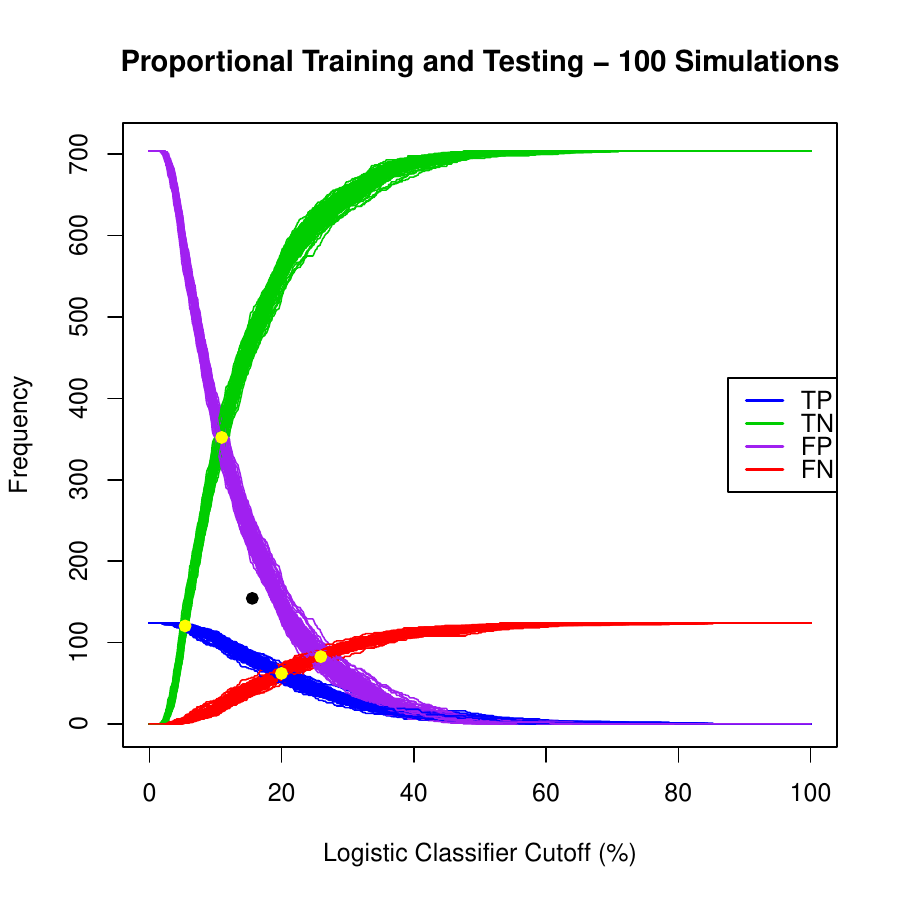}\includegraphics[scale=.22]{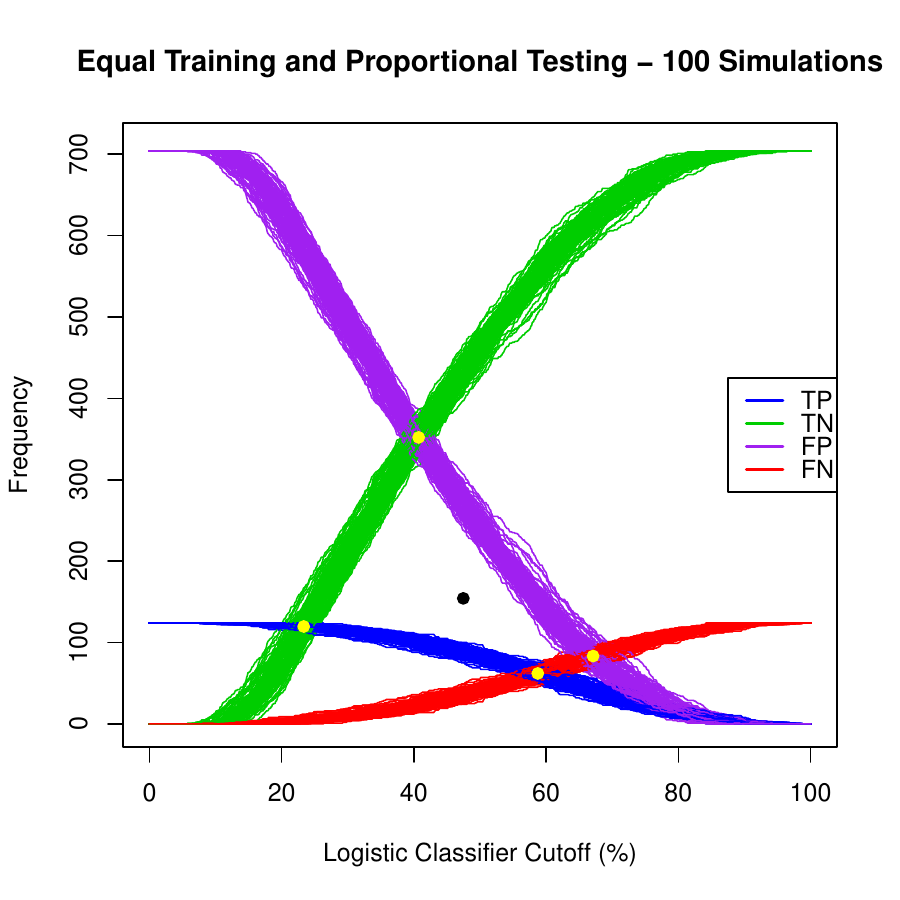}\includegraphics[scale=.22]{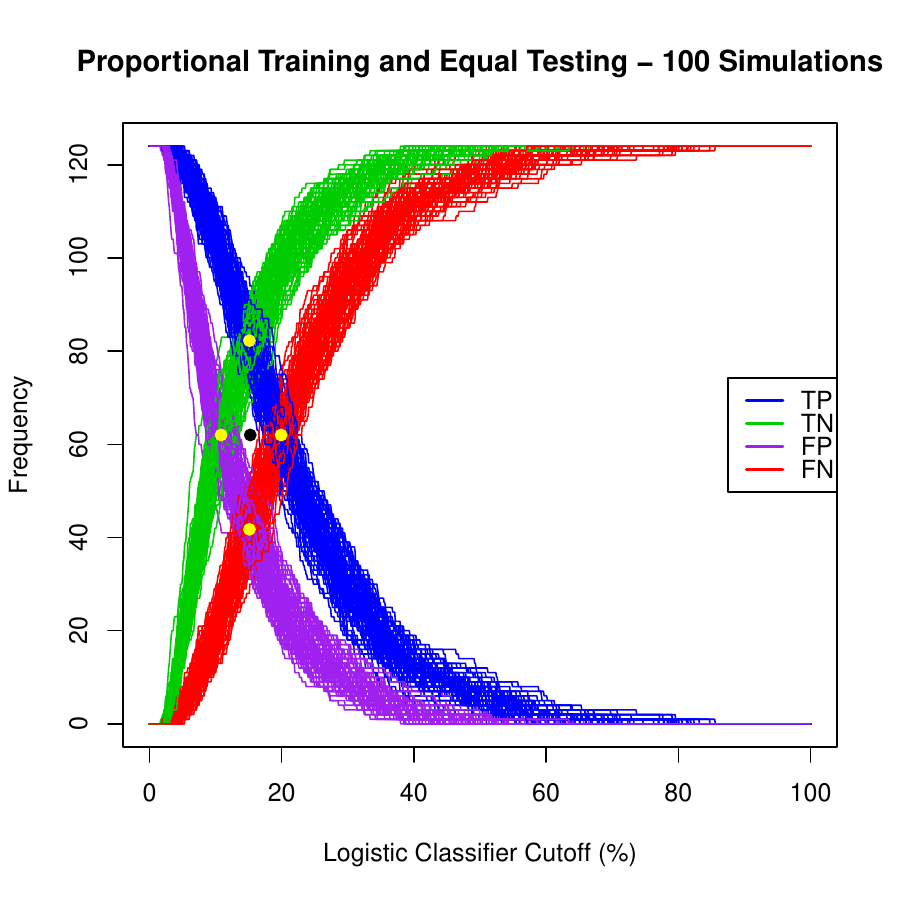}\includegraphics[scale=.22]{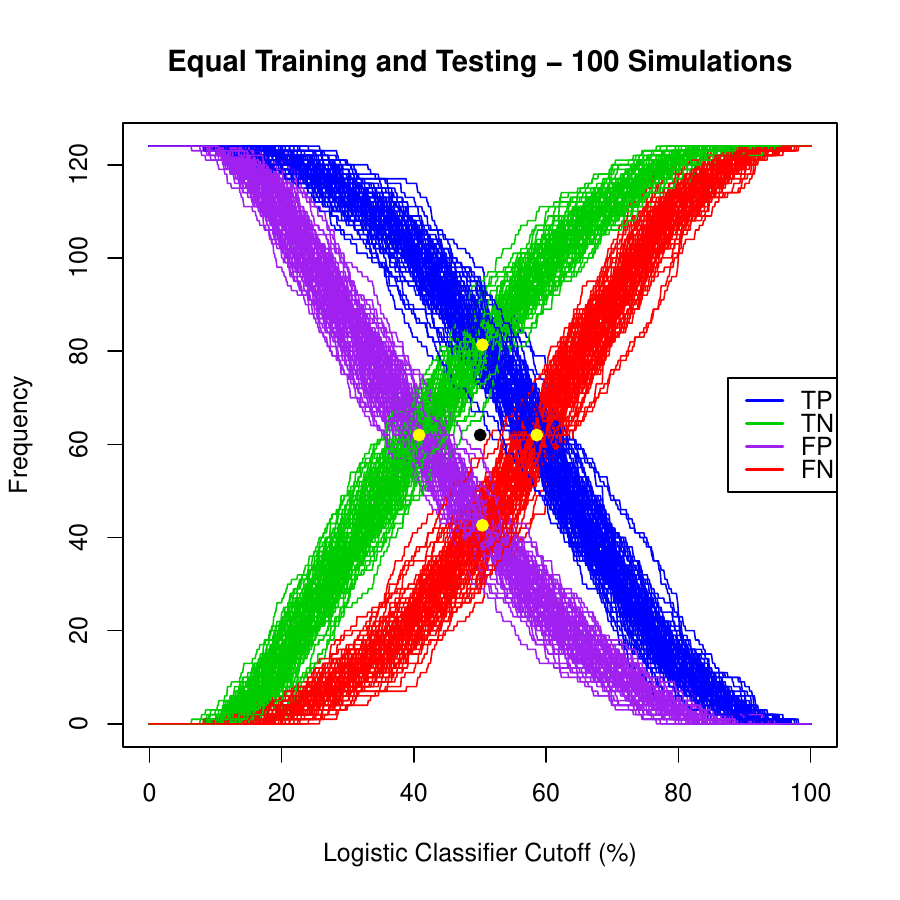}
\includegraphics[scale=.22]{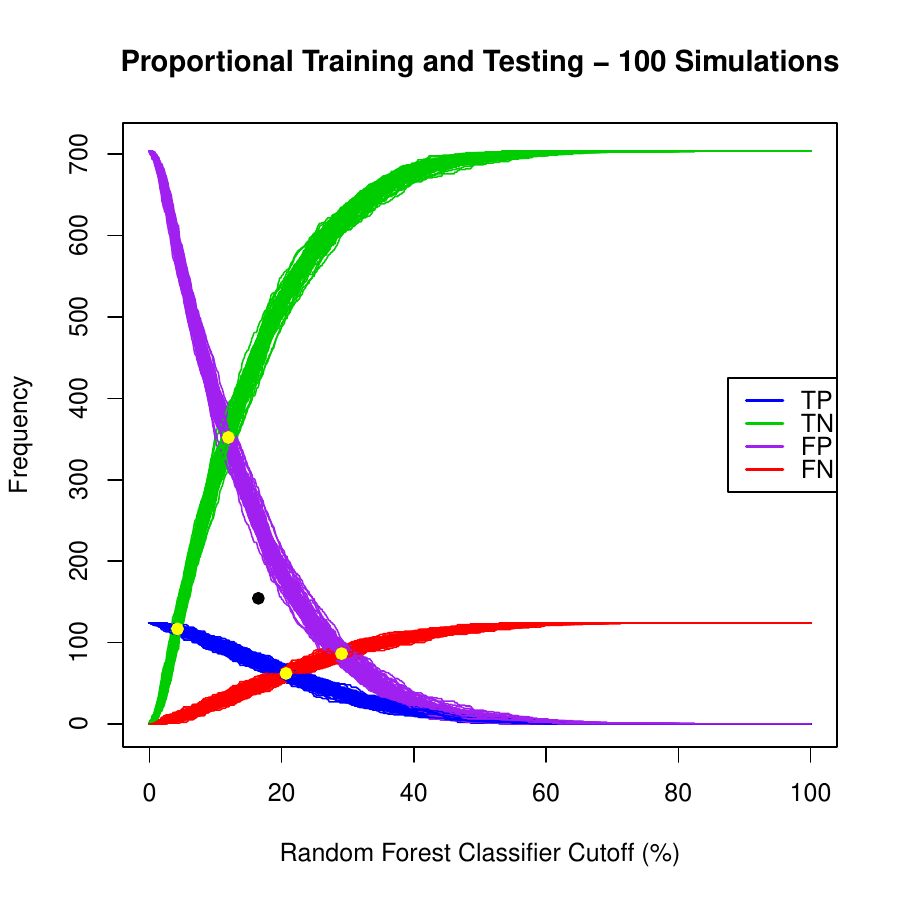}\includegraphics[scale=.22]{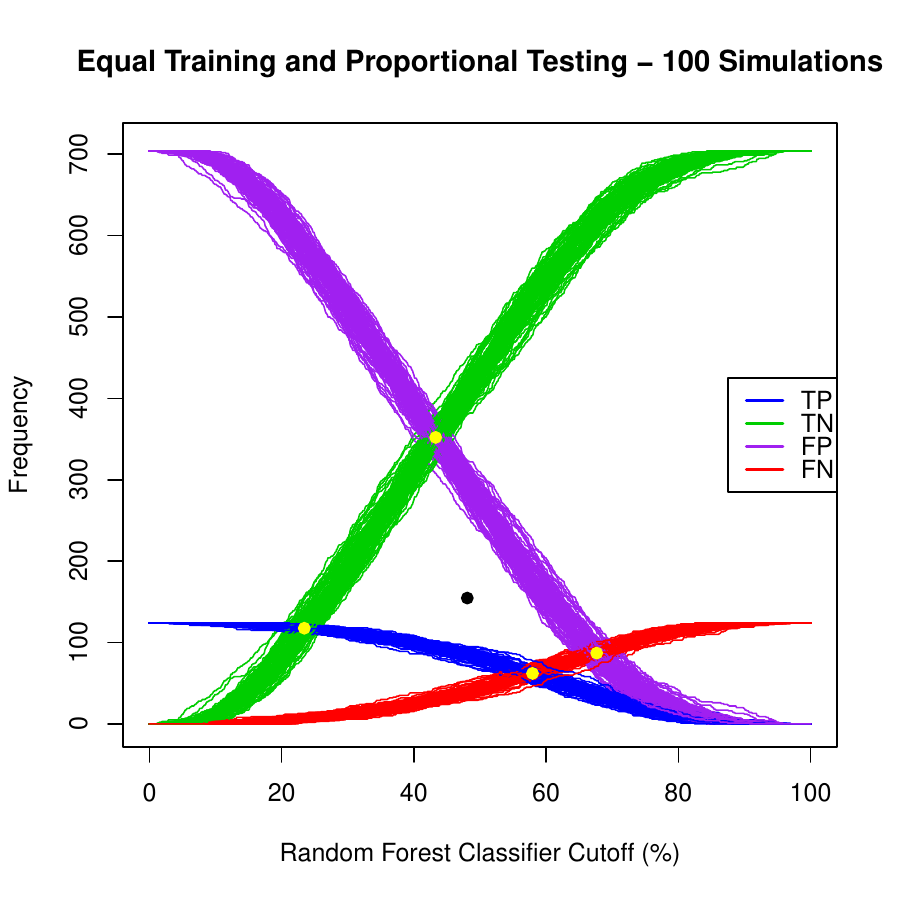}\includegraphics[scale=.22]{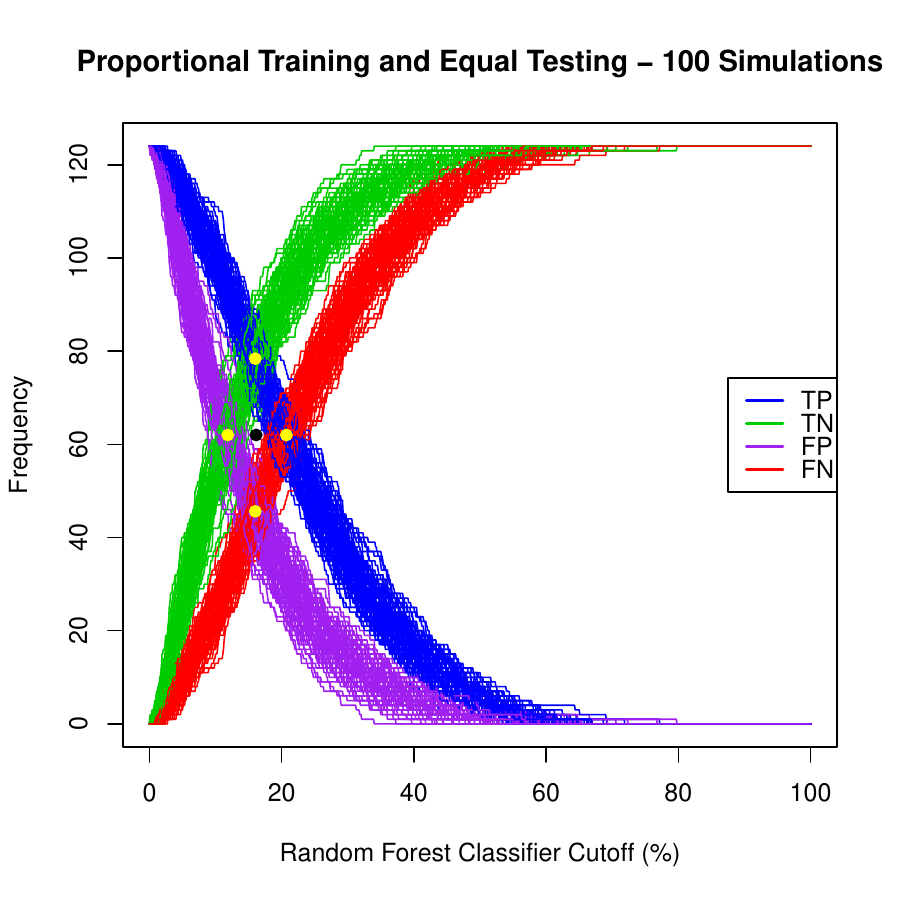}\includegraphics[scale=.22]{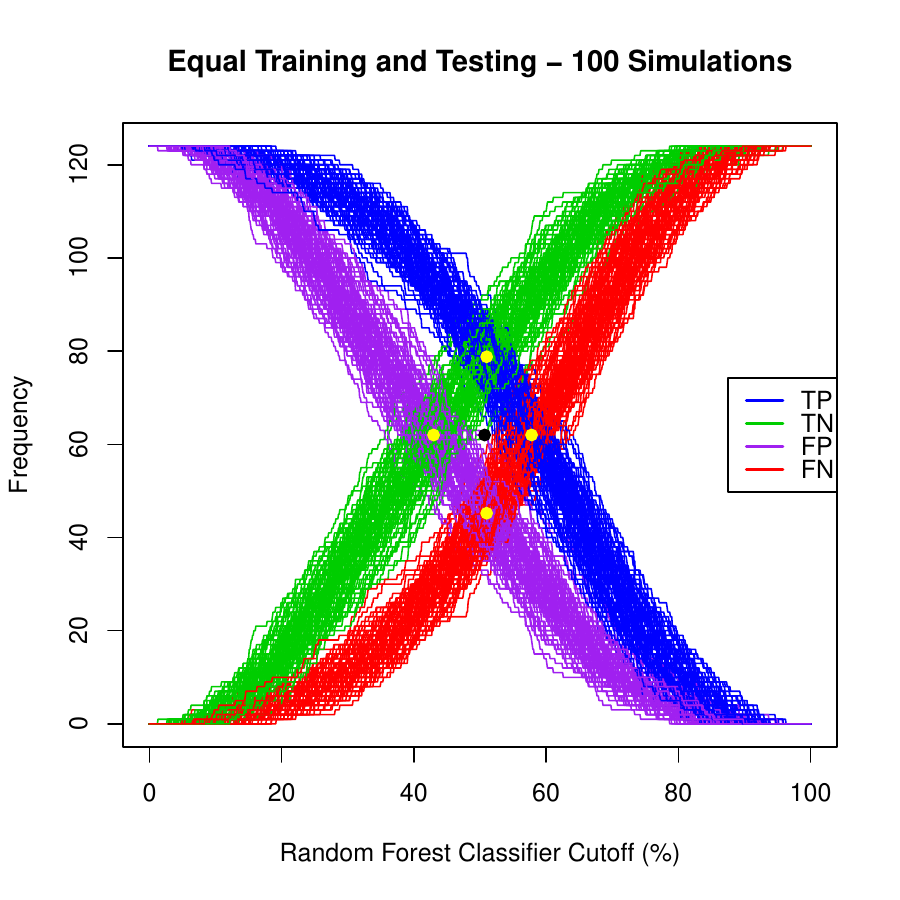}
\caption{Sampling distribution Logistic and Random Forest Classifier Cutoffs for four training/testing scenarios}
\label{glmrf100sim}
\end{center}
\end{figure}

Given a regression model for a dichotomous random variable, e.g., a logistic regression, or random forest, where the output for one tested case is the probability of being a positive case, one can convert such regression model into a classification algorithm by choosing a probability threshold, or a classifier cutoff, and then classify a tested patient as positive if the output of the regression model is greater or equal than the chosen classifier cutoff, and classify a tested patient as negative otherwise. A naive classifier cutoff of 50\% leads to poor prediction performance when the prevalence of the training dataset deviates substantially from 50\%. Indeed, for each of the four training/testing scenarios, the Framingham CHD data, with all seven explanatory variables,  have been split (one simulation) using a training ratio of $\tau = 0.8$. Both logistic regression and random forest models were extracted using the training dataset, and then using both models, all patients in the testing data set have been classified  using successive values of a classifier cutoff ranging from $0\%$ to $100\%$, with a 1\% step, and the number of true/false positive and true/false negative cases have been recorded.  The graphs of the true positive, false positive, false negative and true positive cases as functions of the classifier cutoff for each training/testing scenarios and for both the logistic and random  forest regression models are given in Figure \ref{glmrfOneSim}.  A good classifier cutoff should minimize misclassification. As the classifier cutoff increases, the number of true negative (TN, green curve) increases,  and the number of true positive (TP, blue curve) decreases. Thus an equilibrium classifier cutoff must strike a balance between the number of true positive and true negative cases. One can observe from each of the logistic and random forest graphs that such a balanced classifier cutoff is around 15\% when the training dataset is proportional (first and third training/testing scenarios), and is around 50\% when the training data set is equal (second and fourth training/testing scenarios). In other words, a good and balanced classifier cutoff for both the logistic and random forest regression models appears to be the prevalence of the training dataset, independent from the prevalence of the testing dataset. In light of this observation, we replicated the above simulation one hundred times, and superposed the graphs of each simulation into the same graph, shown in Figure \ref{glmrf100sim}. For these sampling distributions, the yellow dots represent the average point of intersection (centro\"id) of the 100 pairs of curves TN-TP, TP-FN, FN-FP, and TN-FP . The black dot is the average point (centro\"id) of the four yellow points.  For these sampling distributions (100 simulations), the equilibrium classifier cutoff for the logistic regression, i.e., the $x$-coordinate of the black dot, is 15.56\% (resp., 47.48\%, 15.25 and 50.03\%) for the proportional training and testing scenario (resp.., equal training + proportional testing, proportional training + equal testing, and equal training + testing).  Similarly, the equilibrium classifier cutoff for the random forest regression, i.e, the $x$-coordinate of the black dot, is 16.47\% (resp., 48.77\%, 16.14 and 50.70\%) for the proportional training and testing scenario (resp.., equal training + proportional testing, proportional training + equal testing, and equal training + testing).  In light of these sampling distributions, we propose to chose the equilibrium classifier cutoff for both logistic and random forest regression models to be the prevalence of the training dataset.

Note that all above testing for both logistics and random forest regression models were performed using all seven explanatory variables. One can justifiably argue that using all  explanatory variables may not lead to optimal models, and thus one may question the use of such balanced  classifier cutoff. To test this hypothesis, we performed one thousand logistic regression model analyses for each of the four training/testing scenarios, and we counted for each of the intercept and the seven explanatory variables, the number of times such variable was significant using a significance level $\alpha $ of 1\%, 5\% and 10\%. The results of these 1000 simulations are summarized in Table \ref{countingsigvarLogistics}, where $X_0$ stands for the logistic regression intercept. 

\begin{table}[htp]
\begin{center}
\begin{tabular}{cllrrrrrrrr}
\hline
$\alpha$&Training & Testing & $X_0$& $X_1$ &$X_2$ &$X_3$ &$X_4$ &$X_5$ &$X_6$ &$X_7$ \\ \hline
\multirow{4}{*}{1\%} &Proportional & Proportional &  1000 & 1000 & 2 & 1000 & 0 & 0 & 0 &1000   \\
&Equal & Proportional &1000 & 1000 & 21 & 622 & 0 &5 &0 &1000   \\
&Proportional & Equal &  1000 & 1000 & 4 & 1000 & 0 & 0 & 0 & 1000  \\
&Equal & Equal &  1000 & 1000 & 21 & 656 & 2 & 7 & 1 & 1000  \\
\hline
\multirow{4}{*}{5\%} &Proportional & Proportional & 1000 & 1000 & 82 & 1000 & 0 & 25 & 2 & 1000  \\
&Equal & Proportional & 1000 & 1000 & 125 & 887 & 15 & 69 & 19 & 1000 \\
&Proportional & Equal & 1000 & 1000 & 71 & 1000 & 0 & 18 & 0 & 1000  \\
&Equal & Equal &  1000 & 1000 & 130 & 883 & 15 & 62 & 15 & 1000 \\
\hline
\multirow{4}{*}{10\%} &Proportional & Proportional & 1000 & 1000 & 185 & 1000 & 1 & 73 & 12 & 1000 \\
&Equal & Proportional &  1000 & 1000 & 257 & 953 & 43 & 150 & 18 & 1000 \\
&Proportional & Equal & 1000 &  1000 & 207 & 1000 & 1 & 56 & 6 & 1000  \\
&Equal & Equal &  1000 & 1000 & 228 & 941 & 43 & 124 & 32 & 1000  \\
\hline
\end{tabular}
\end{center}
\caption{Logistic regression variable significance analysis for four training/testing scenarios - 1000 simulations}
\label{countingsigvarLogistics}
\end{table}%

For all three significance levels $\alpha$, the $y$-intercept, age $X_1$, and number of cigarettes smoked per day $X_7$ were statistically significant 1000 time (out of 1000). Moreover, systolic blood pressure $X_3$ were statistically significant 1000 times when the testing dataset were proportional (15\% prevalence),  and between 622 and 953 times when the testing dataset were equal (50\% prevalence). For a significance level of $\alpha = 10\%$, the total cholesterol level $X_2$ were statistically significant between 185 to 257 times (out of 1000), which is roughly speaking between 20 to 25\% of the 1000 simulation runs. In light of this logistic regression variable significance analysis, for each of the four training/testing scenarios and for the same data split (training ratio $\tau = 0.8$), we performed three logistic regression models: Logistic Model 1 with all seven explanatory variables, Logistic Model 2 with only age $X_1$, systolic blood pressure $X_3$ and number of cigarettes smoked per day $X_7$, and Logistic Model 3 with the total cholesterol  $X_2$ added to the three variables in model 2. For a given training/testing scenario and for the same data split,  the three logistic regression models were converted into three logistic classifiers using a balanced/equilibrium classifier cutoff equal to the prevalence of the training dataset, and the number of true positive cases (out of 124 patient tests for each training/testing scenarios) were recorded. This process was repeated 1000 times, and the average number of true positive predictions for each of the three regression classification models  are summarized in Table \ref{TP3glmmodels}. 

\begin{table}[htp]
\begin{center}
\begin{tabular}{llccc}
\hline
\multirow{2}{*}{Training} & \multirow{2}{*}{Testing} & Logistic 1  & Logistic 2  & Logistic 3 \\ 
& & $(X_1, \dots, X_7)$ & $(X_1, X3, X_7)$ &  $(X_1, X2, X3, X_7)$\\
\hline
Proportional & Proportional & 81.80 & 81.81 & 82.59\\
Equal & Proportional & 82.68 & 82.80 & 83.06 \\
Proportional & Equal & 81.85&  82.80& 82.67\\
Equal & Equal & 82.18 & 81.77 & 82.73\\
\hline
\end{tabular}
\end{center}
\caption{Average number of true positive cases (out of 124) for three Logistic classification models across four training/testing scenarios - 1000 simulations}
\label{TP3glmmodels}
\end{table}%
Table \ref{TP3glmmodels}. shows that the number of true positive predictions is roughly the same regardless of what variables are used for the logistic regression. In other words, while different choice of variables may make a difference for the logistic regression where the outcome is a probability, such variable choice seems to not be relevant for a logistic classifier across all four training/testing scenarios with an (equilibrium) classifier cutoff equal to the prevalence of the training dataset. A modeler can thus either use all variables, or run few simulation runs and use the subset of variables that are statistically significant for the logistic regression. Finally, comparing the average number of true negative, false positive and false negative were consisting with the results in Table \ref{TP3glmmodels}, and thus we decided to not report the average frequencies. 

\subsection{Training Ratio Analysis and Classification Algorithm Comparison}\label{comparealgo}
We consider in this subsection the following eight classification algorithms, and we compare their predictive performances  using the Framingham CHD data across the four training/testing scenarios.  Note that the last two classification algorithms, i.e., the double discriminant scoring of type 1 and type 2, are introduced in this paper, and have not been considered yet in the literature.

\begin{enumerate}
\item The Extreme Gradient Boosting (XGB), for which the outcomes for one tested case are the two probabilities of being a positive and negative case respectively. Therefore, we assign a tested case to Group 1 (having CHD) if the probability of being positive is greater than the probability of being negative, otherwise we assign the tested case to Group 2 (not having CHD),
\item Support Vector Machine (SVM),
\item Random Forest Classifier (RF), where the classifier cutoff is set to be the prevalence of the testing dataset (as shown in subsection \ref{LandRF}),
\item Logistic Classifier (Logit), where the classifier cutoff is set to be the prevalence of the testing dataset (as shown in subsection \ref{LandRF}), and the link function is the logit function,
\item Linear Discriminant Function (LD), 
\item Quadratic Discriminat Function (QD),
\item Double Discriminant Scoring of Type 1 (DDS1): a tested patient is assigned to Group 1 (having CHD) if either the linear or quadratic discriminant models assign the tested patient to Group 1, otherwise, the tested patient is assigned to Group 2 (not having CHD), 
\item Double Discriminant Scoring of Type 2 (DDS2): a tested patient is assigned to Group 1 (having CHD) if both the linear and quadratic discriminant  models assign the tested patient to Group 1,  otherwise, the tested patient is assigned to Group 2 (not having CHD).  
\end{enumerate}

\begin{figure}[htbp]
\begin{center}
\includegraphics[scale=.3]{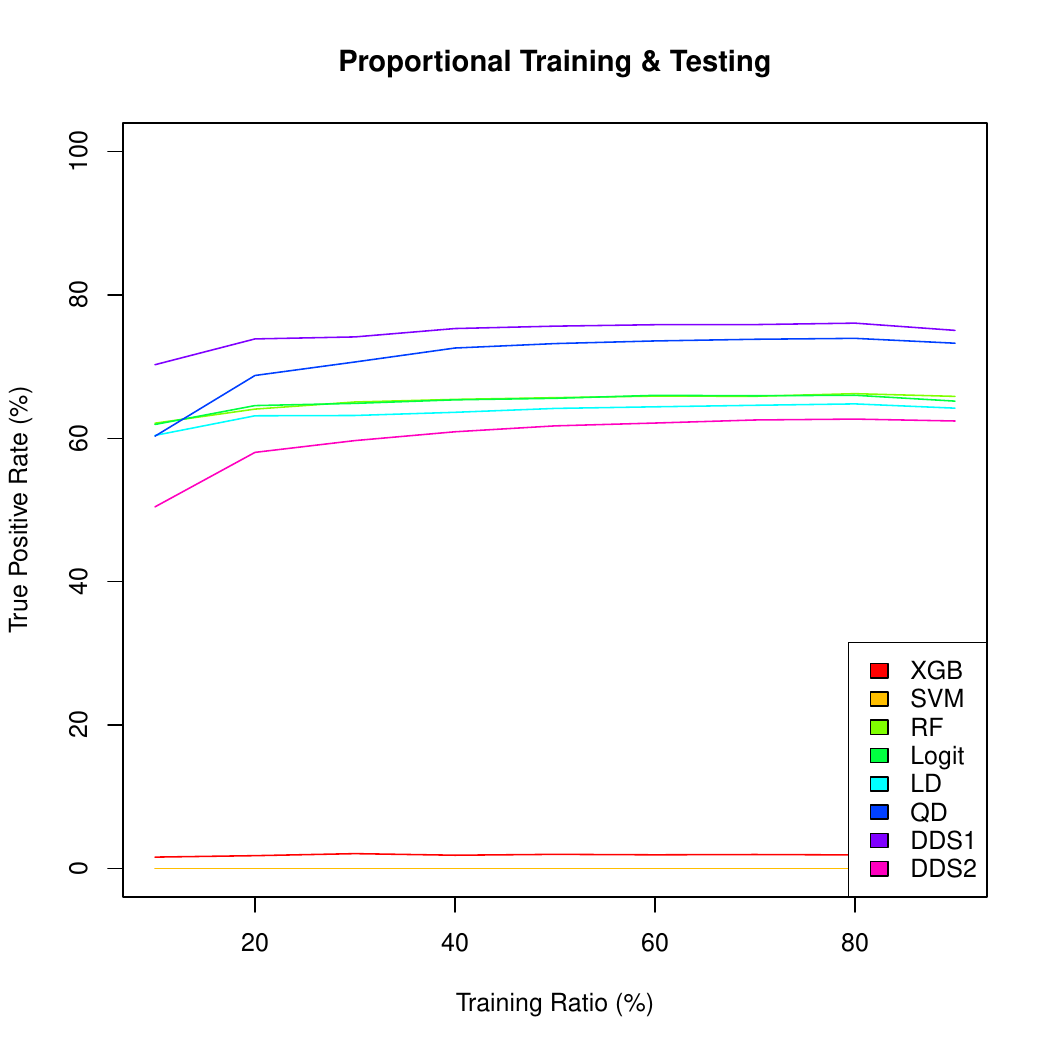}\includegraphics[scale=.3] {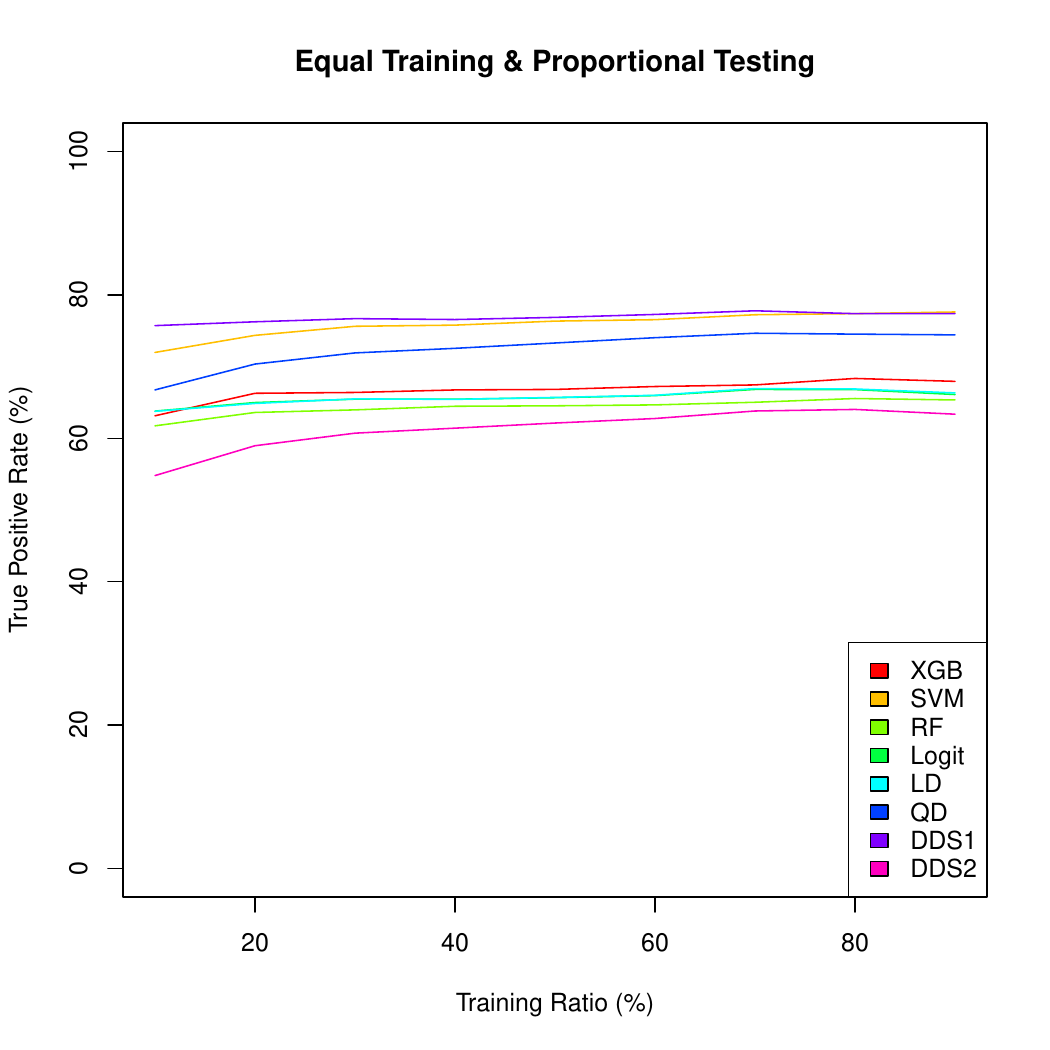} \includegraphics[scale=.3]{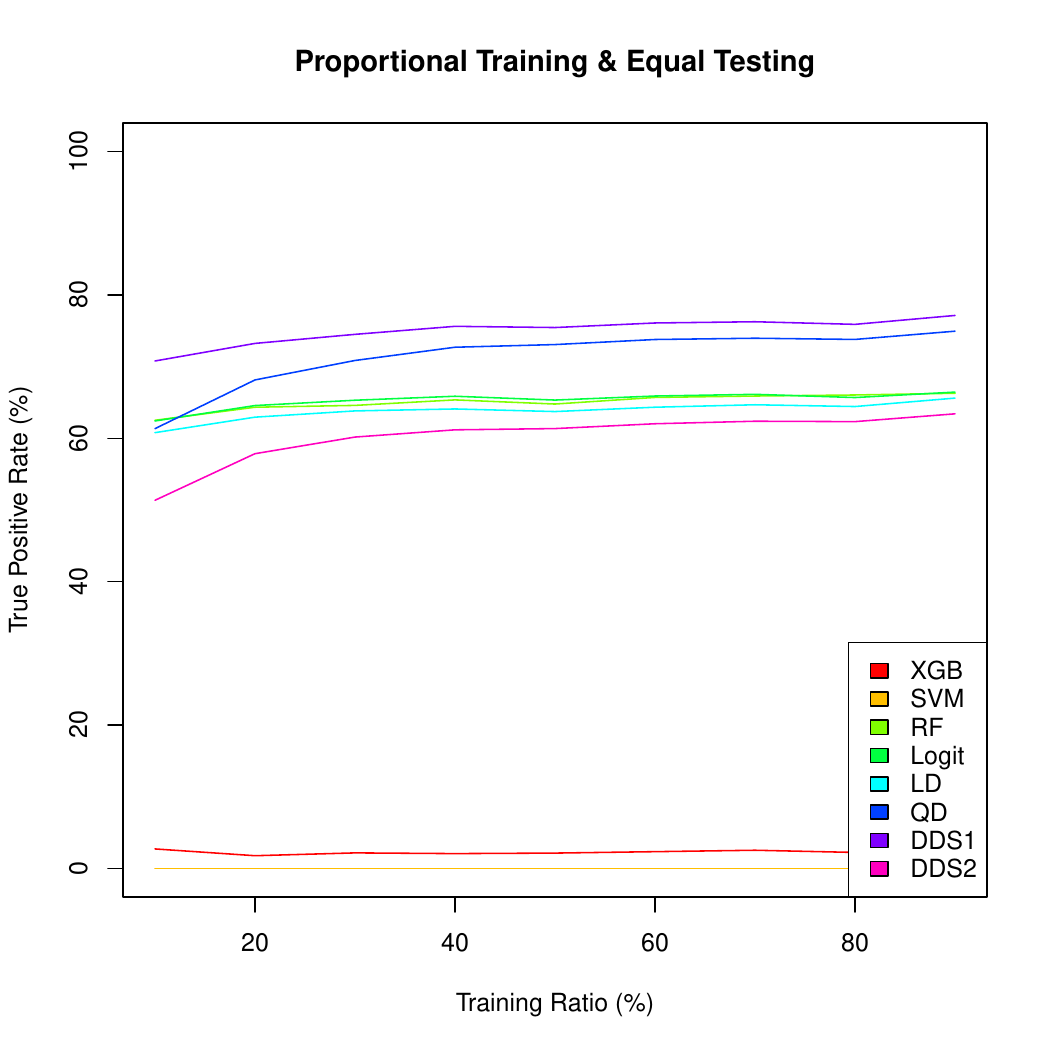}\includegraphics[scale=.3]{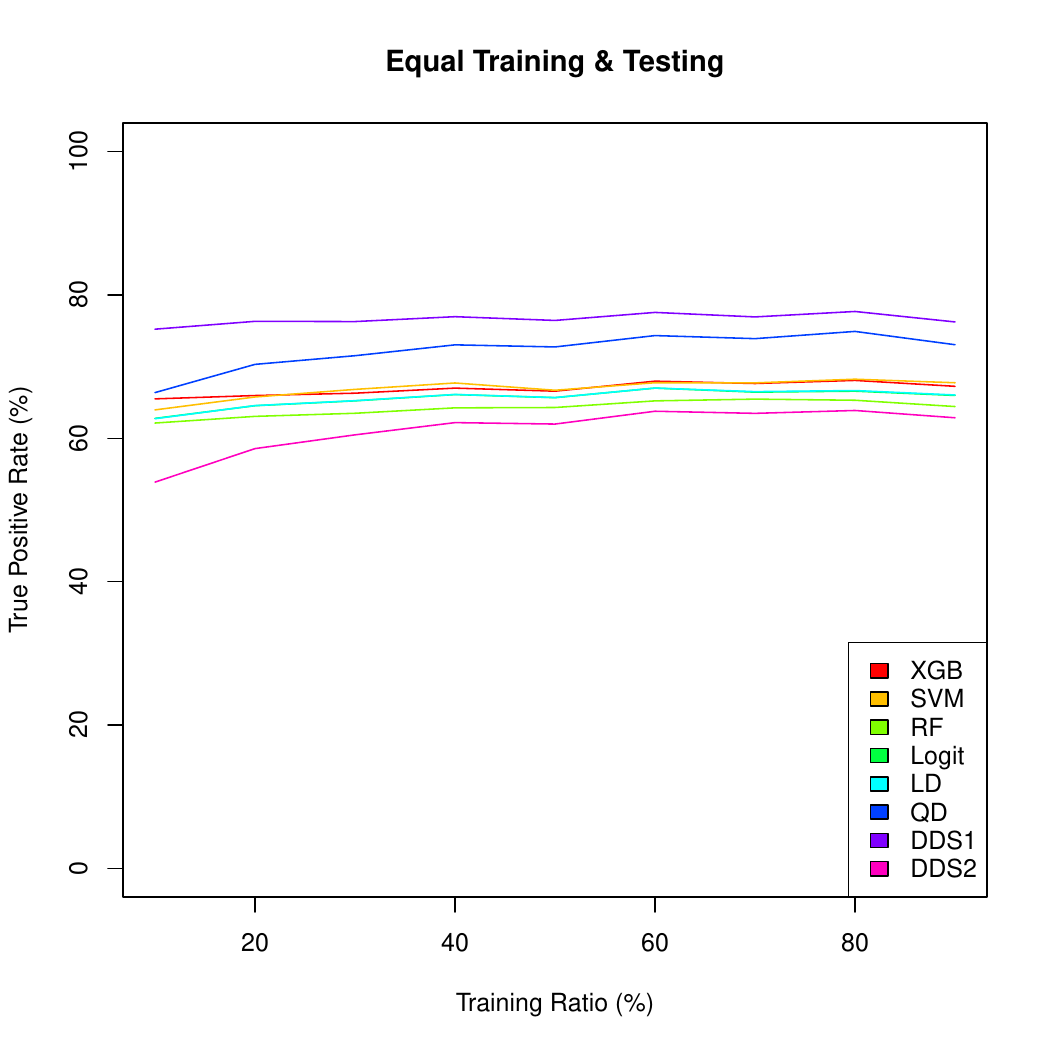}
\caption{Mean of 100 True Positive Rates for eight classification algorithms as functions of the training ratio across four training/testing scenarios}
\label{TS8algoTPR}
\end{center}
\end{figure}

Note that both the linear and quadratic discriminant functions \cite{DAbook} in this research assume normality, and are derived as the difference of the log-likelihood functions. When the  (multi-variate) population variances of Groups 1 and 2  are assumed to be equal (resp., different), the maximum log-likelihood discriminant rule leads to the linear  (resp., quadratic) discriminant function. We made the choice to simply consider both instances without running a Bartlett test.  One can think of the Double Discriminant Scoring of Type 1 (resp., Type 2) as a ``liberal'' (resp., ``conservative") combination of the linear and quadratic discriminant functions. We compare the mean predictive performance of these eight classification algorithms for training ratio $\tau = 0.1, 0.2, \dots, 0.8, 0.9$ across four training/testing scenarios using 100 simulations for each classification algorithm. All predictive performance metrics in the classification performance matrix, Table \ref{classperfmatrix}, i.e., TP, FP, FN, TN, TPR, TNR, PPrec, NPrec, OPrev, EPrev and Acc, have been recorded, and the mean of 100 simulations have been computed for each one of the four training/testing scenarios. In Figure \ref{TS8algoTPR}, the graphs of the true positive rates as functions of the training ratio $\tau$ are plotted,  and lead to the following remarks. First, both Extreme Gradient Boosting and Support Vector Machine classification algorithms perform horribly when the training dataset are proportional, that is, when the prevalence of the training dataset is equal to the prevalence of the Framingham CHD data (15\%).  These two algorithms are extremely biased toward Group 2, as they predicted almost all tested patients as not having CHD when the training dataset were proportional. However, they performed better when the training dataset were equal, with the Support Vector Machine being the second best algorithm when the training dataset is equal and the testing dataset is proportional. Second, the Double Discriminant Scoring  of Type 1 consistently outperformed all other algorithms for all training data ratios and across all four training/testing scenarios. Moreover, the true positive rates for the Double Discriminant Scoring of Type 1 are fairly constant across all four training/testing scenarios when the training ratio $\tau$ was higher than 0.4, leading to think that a training data set of size 250 or higher leads to consistent predictions. This generalizability finding is very important for predictions in the health sector  as it would enable a modeler to be confident about applying an optimal model from one dataset to another dataset from a different geographical area with a different distribution and/or prevalence. 


In Table \ref{Perf8algo8020}, the mean accuracy, true positive rate, true negative rate, number of true positive, false negative, false positive and true negative tested patients  (out of 100 simulations) using a training ratio of $\tau = 0.8$ for each of the eight classification algorithms are summarized. Using accuracy as the only metric for assessing the predictive performance of a classification model is dangerous. Indeed, the highest mean accuracy is 84.95\% which occurred for the Extreme Gradient Boosting and the Support Vector Machine when both the training and testing datasets were proportional. However, the corresponding true positive rates are 1.98\% and 0\% respectively, as both models predicted (almost) all the 828 test patients as negative. As mentioned previously, both Extreme Gradient Boosting and Support Vector Machine are biased and flawed when the training dataset is proportional, regardless of the prevalence of the testing dataset. Let's focus our attention to the true positive rates.  Out of the 32 sampling distribution (eight algorithms and four training/testing scenarios), only five true positive rates are above 75\%. One of these is the Support Vector Machine when the training dataset is equal and the testing dataset is proportional, and the four other  are the Double Discriminant Scoring of Type 1 for all four training/testing scenarios. While the Support Vector Machine's mean true positive rate is 76.44\% (the third highest) for an equal training and proportional testing, its mean true positive rate drops to 68.04\% when testing is equal, and in addition for having a null true positive rates when the training dataset is proportional, we believe that the Support Vector Machine, while performing well with an equal training dataset, is sensitive to the prevalence of the testing dataset and thus one can't be confident about using the model for patients of unknown CHD status even when making sure that the prevalence of the training dataset remains at 50\%. However, the mean true positive rates for Double Discriminant Scoring of Type 1 for all four training scenarios are higher than 75\%, which means that this classification algorithm performs consistently well for all training/testing scenarios, and as shown in Figure \ref{TS8algoTPR}, it performs consistently well whenever the size of the training dataset is above 250 observations. This shows that the Double Discriminant Scoring of Type 1 is generalizable.  Note that the cost to be paid by the Double Discriminant Scoring of Type 1 classification method for having the highest true positive rate is not having the highest true negative rates. This being said, reliably predicting true positive patients and minimizing false positive is crucial  in medicine and public health. 
\begin{table}[htp]
\begin{center}
\begin{tabular}{ccrrrrrrrr}
\hline
Algo&Training/Testing &  Acc\% & TPR\% & TNR\%  & TP & FN & FP & TN  \\ \hline
\multirow{4}{*}{XGB} &
Prop./Prop. &  84.95 & \textcolor{red}{1.98} & 99.57 & 2.46 & 121.54 & 3.02 & 700.94   \\
&Equal/Prop. & 61.94 & 66.98 & 61.05 & 83.05 & 40.95 & 274.22 & 429.78  \\
&Prop./Equal & 50.83  & \textcolor{red}{2.18} & 99.49 & 2.70 & 121.30 & 0.63 & 123.37    \\
&Equal/Equal &  64.86 & 67.57  & 62.15 & 83.79 & 40.21 & 46.93 & 77.07 \\
\hline
\multirow{4}{*}{SVM} &
Prop./Prop. &  84.95 & \textcolor{red}{0.00} &  100.00& 0.00& 124 &0.00  &704    \\
&Equal/Prop. & 57.95  & \textbf{76.44}  & 54.70 & 94.78& 29.22 & 318.94 &385.06    \\
&Prop./Equal &  50.00 & \textcolor{red}{0.00} & 100.00 & 0.00&  124.00&0.00  & 124.00     \\
&Equal/Equal &  66.41 & 68.04 & 64.77 & 84.37& 39.63 & 43.68 &80.32   \\
\hline
\multirow{4}{*}{RF} &
Prop./Prop. & 60.91  &65.41  & 60.12 & 81.11&  42.89& 280.78 & 423.22   \\
&Equal/Prop. & 62.11  & 64.89 & 61.62 & 80.46& 43.54 & 270.22 & 433.78   \\
&Prop./Equal &  62.87 & 65.66 & 60.08 & 81.42&  42.58& 49.50 & 74.50     \\
&Equal/Equal & 63.59  & 65.20 & 61.97 & 80.85& 43.15 & 47.16 & 76.84  \\
\hline\multirow{4}{*}{Logit} &
Prop./Prop. &65.77   & 65.41 & 65.77 & 81.62&  42.38& 241.01  &462.99    \\
&Equal/Prop. &  65.68 & 65.60 &  65.69& 81.34& 42.66 & 241.53 & 462.47   \\
&Prop./Equal &  66.12 & 65.99 & 66.25  & 81.83 & 42.17 & 41.85 & 82.15     \\
&Equal/Equal & 66.21  & 66.37 & 66.07 & 82.30&  41.70& 42.09 & 81.91  \\
\hline\multirow{4}{*}{LD} &
Prop./Prop. &  66.48 & 64.57 &66.81  & 80.07&  43.93&  233.65&  470.35  \\
&Equal/Prop. & 65.71  & 65.81 & 65.69 & 81.34&  42.40& 241.51 & 462.49   \\
&Prop./Equal & 65.98  & 64.73 & 67.23 & 80.26&  43.74&  40.64& 83.36     \\
&Equal/Equal &  66.30 &  66.60& 65.99 & 82.58& 41.42 & 42.17 & 81.83  \\
\hline\multirow{4}{*}{QD} &
Prop./Prop. &  59.18 & 74.51 & 56.48 & 92.39 & 31.61 & 306.36 & 397.64   \\
&Equal/Prop. &  58.87 & 73.54 & 56.29 & 91.19& 32.81 & 307.73 & 396.27   \\
&Prop./Equal & 65.13  & 73.87 & 56.39 & 91.60& 32.40 & 54.08 & 69.92     \\
&Equal/Equal & 65.79  & 74.68 & 56.90 & 92.60& 31.40 & 53.45 & 70.55  \\
\hline\multirow{4}{*}{\textbf{DDS1}} &
Prop./Prop. &  58.21 & \textbf{76.53} &  54.48& 94.90& 29.10 & 316.96 & 387.04   \\
&Equal/Prop. & 57.49  & \textbf{76.86} &54.08 &  95.31& 28.69&  323.30& 380.70    \\
&Prop./Equal &  65.41 &  \textbf{75.94}&  54.87& 94.17&  29.83& 55.96 & 68.04      \\
&Equal/Equal & 65.87  & \textbf{77.12} & 54.62 & 95.63&  28.37& 56.27 & 67.73  \\
\hline\multirow{4}{*}{DDS2} &
Prop./Prop. &  67.45 & 62.55 & 68.32 & 77.56&  46.44&  223.05&  480.95  \\
&Equal/Prop. &  67.09  & 62.48  & 67.91  & 77.48 & 46.52 & 225.94 & 478.06   \\
&Prop./Equal &  65.70 & 62.65  & 68.74 & 77.69& 46.31 & 38.76 & 85.24     \\
&Equal/Equal &  66.21 & 64.15 & 68.27 & 79.55& 44.45 & 39.35 & 84.65  \\
\hline
\end{tabular}
\end{center}
\caption{Mean prediction performance metrics for eight classification algorithms across four training/testing scenarios, 100 simulations,  for a training ratio of $\tau = 0.8$}
\label{Perf8algo8020}
\end{table}%

\subsection{Double Discriminant Scoring Methodology}\label{ddsm}
The Double Discriminant Scoring of Type 1 consistently performed the best across all four training/testing scenarios and for all training ratio $\tau = 0.1, 0.2, \dots, 0.9$ when comparing true positive rates using all seven explanatory variables. In this subsection, we establish a methodology to not only determine the optimal variable selection for a machine learning classification algorithm, but also to derive a hierarchy for the optimal subset of explanatory variables.  We illustrate this methodology using the Framingham CHD. Given a multivariate data with $p$ explanatory variables $(X_1, \dots, X_p)$ and one response variable $Y$, a machine learning classification algorithm can be derived using either all or a subset of these $p$ explanatory variables, thus $2^p-1$ possible models for the same classification algorithm can be derived. For the Framingham CHD data, there are $2^7-1 = 127$ possible subsets of the seven variables that can be used for the classification algorithm. Using a training ratio $\tau = 0.8$, and for each of one of the four training/testing scenarios, and for each of the 127 possible variable selections, we performed 1000 prediction simulations using the Double Discriminant Scoring of Type 1 algorithm, that is, randomly split the data into training and testing dataset, train the model using the training data set, classify observations in the testing data set into Group 1 (positive, having CHD) and Group 2 (negative, not having CHD), compute the classification performance matrix (Table \ref{classperfmatrix}), and compute the mean of these 1000 prediction simulation runs for each performance metric in the classification performance matrix. Therefore, one can derive a data frame  with $2^p-1 = 127$ rows (number of variable sub-selections) and as many columns as the number of considered performance metrics, and sort this data frame with respect to a prediction performance metric (column), e.g., the true positive rate. We performed the above analysis using a paired design for the Linear Discriminant function (LD), the Quadratic Discriminant function (QD), the Double Discriminant Scoring of Type 1 and the Double Determinant Scoring of Type 2. However, we focus our attention to the Double Discriminant Scoring of Type 1. Using the mean  true positive rate (out of 1000 simulations), the data frame of 127 mean true positive rates have been sorted from the highest to the lowest true positive rate for each of the four training scenarios, and the top five variable selections are reported in Table \ref{Top6DDSTPR}. Note that $(1,2,7)$ in Table \ref{Top6DDSTPR}. refers to $(X_1, X_2, X_7)$ and hence, to the model using age $X_1$, total cholesterol $X_2$ and number of cigarettes smoked per day $X_7$ for the predictions.  
\begin{table}[htp]
\begin{center}
\begin{tabular}{cclc}
\hline
Training & Testing  & Variables & Mean True Positive Rate \%  \\ \hline
\multirow{6}{*}{Proportional} & \multirow{6}{*}{Proportional} & (1,2,4,5,6,7)  &78.04391\\
&&(1,2,4,5,7)    &77.60114 \\
&&(1,2,4,6,7)    &76.96811\\
&&(1,2,4,7)     &76.30827 \\
&&(1,2,3,4,5,6,7) &76.18401 \\
&&\vdots   &\vdots \\
\hline
\multirow{6}{*}{Equal} & \multirow{6}{*}{Proportional} & 
(1,2,4,5,6,7) &  78.29709 \\
&&(1,2,4,5,7)    &77.43829 \\
&&(1,2,3,4,5,6,7) & 77.42931 \\
&&(1,2,4,6,7)    &77.13655 \\
&&(1,2,3,4,5,7)   &76.82127 \\
&&\vdots   &\vdots \\
\hline
\multirow{6}{*}{Proportional} & \multirow{6}{*}{Equal} & 
(1,2,4,5,6,7)  &78.06162 \\
&&(1,2,4,5,7)    &77.54947 \\
&&(1,2,4,6,7)    &77.00361 \\
&&(1,2,4,7)     &76.43153 \\
&&(1,2,3,4,5,6,7)  &76.20655 \\
&&\vdots   &\vdots \\
\hline
\multirow{6}{*}{Equal} & \multirow{6}{*}{Equal} & 
(1,2,4,5,6,7) & 78.25840 \\
&&(1,2,4,5,7)   & 77.68898 \\
&&(1,2,4,6,7)   & 76.90521 \\
&&(1,2,4,7)    & 76.53085 \\
&&(1,2,3,4,5,6,7) & 76.44378\\
&&\vdots   &\vdots \\
\hline
\end{tabular}
\end{center}
\caption{The top five variable sub-selections with respect to the true positive rates for the double discriminant scoring of type 1 across all four training/testing scenarios.}
\label{Top6DDSTPR}
\end{table}%

From Table \ref{Top6DDSTPR}., on average, using all explanatory variables except systolic blood pressure $X_3$ led to the highest true positive rates ($78\%$) for all four training/testing scenarios. Note that the model using all seven explanatory variables, analyzed in subsection \ref{comparealgo}, appears in the top five for each of the four training/testing scenarios and has a true positive rates between 76\% and 77\%.  Further analyzing the ranking of all 127 variable sub-selections with respect to the true positive rates shows that the true positive rate ranking satisfies the Bellman principal for optimality, that is, ``\emph{an optimal policy has the property that whatever the initial state and initial decision are, the remaining decisions must constitute an optimal policy with regard to the state resulting from the first decision}''  \cite{BellmanBook}. Note that out of the 127 variable sub-selections, 7 (resp.., 21, 35, 35, 21, 7 and 1) are of size 1 (resp., 2, 3, 4, 5, 6 and 7) variables. In Table \ref{VarHiearchyTable}., the optimal variable sub-selection per number of variables and the corresponding mean true positive rates are given for each of the four training/testing scenarios, and shows that the same optimal variable hierarchy, age $X_1$, diastolic blood pressure $X_4$, number of smoked cigarettes per day $X_7$, total cholesterol $X_2$, BMI $X_5$ and heart rata $X_6$, stands for all four training/testing scenarios. 


\begin{table}[htp]
\begin{center}
\begin{tabular}{cllllc}
\hline
Number of & \multirow{2}{*}{Training} & \multirow{2}{*}{Testing}   &  Optimal  & Mean \%  \\
Variables&&& Variable Hierarchy &  True Positive Rate \\\hline
1 & \multirow{6}{*}{Proportional} & \multirow{6}{*}{Proportional} &
(1)  & 62.94364\\
2&&&(1,4)    & 69.54167 \\
3&&&(1,4,7)    & 72.64327\\
4&&&(1,4,7,2)     & 76.30827 \\
5&&&(1,4,7,2,5) & 77.60114 \\
6&&&\textbf{(1,4,7,2,5,6)}  & \textbf{78.04391} \\
7&&&(1,4,7,2,5,6,3)  & 76.18401 \\
\hline
1 & \multirow{6}{*}{Equal} & \multirow{6}{*}{Proportional} & 
 (1)  & 63.55076\\
2&&&(1,4)    & 69.67313\\
3&&&(1,4,7)    & 72.93037\\
4&&&(1,4,7,2)     & 76.34774 \\
5&&&(1,4,7,2,5) & 77.43829 \\
6&&&\textbf{(1,4,7,2,5,6)}  & \textbf{78.29709} \\
7&&&(1,4,7,2,5,6,3)  &77.42931 \\
\hline
1 & \multirow{6}{*}{Proportional} & \multirow{6}{*}{Equal} & 
 (1)  & 63.14595\\
2&&&(1,4)    & 69.61105\\
3&&&(1,4,7)    & 72.78043\\
4&&&(1,4,7,2)     & 76.43153 \\
5&&&(1,4,7,2,5) & 77.54947 \\
6&&&\textbf{(1,4,7,2,5,6)}  & \textbf{78.06162} \\
7&&&(1,4,7,2,5,6,3)  & 76.20655\\
\hline
1 & \multirow{6}{*}{Equal} & \multirow{6}{*}{Equal} & 
 (1)  & 63.04836\\
2&&&(1,4)    & 69.87731\\
3&&&(1,4,7)    & 73.12409\\
4&&&(1,4,7,2)     & 76.53085 \\
5&&&(1,4,7,2,5) & 77.68898\\
6&&&\textbf{(1,4,7,2,5,6)}  &  \textbf{78.25840}\\
7&&&(1,4,7,2,5,6,3)  & 76.44378 \\
\hline
\end{tabular}
\end{center}
\caption{Optimal variable hierarchy per number of variable for the Double Discriminant Scoring of Type 1 across four training/testing scenarios.}
\label{VarHiearchyTable}
\end{table}%

A similar optimal variable hierarchy analysis with respect to the true negative rates leads to systolic blood pressure $X_3$ as being the optimal variable sub-selection across all four training/testing scenarios. We conjecture that the optimal variable hierarchy with respect to the true negative rate is always the complement of the optimal variable hierarchy with respect to the true positive rate. Additionally, the optimal variable hierarchy analysis with respect to the positive precision leads to the same variable hierarchy $(X_1, X_4, X_7, X_2, X_5, X_6)$ as for the true positive rate across the four training/testing scenarios, with a 93\% positive precision when the testing is proportional, and a 71\% positive precision when the testing is equal. The dependency of the positive precision on the prevalence of the testing dataset is expected, but the high positive precision when the testing dataset is proportional is agreeably surprising. 

In light of the above analysis, we believe that performing $2^p-1$ sampling distributions for a multivariate data with $p$ explanatory variables is not necessary to determine the optimal variable hierarchy with respect to the true positive and negative rates. One can perform at most $p(p+1)/2$ of such tests. For instance, for the Framingham CHD data, the seven sampling distributions of single variables  with respect to the true positive rate shows that age $X_1$ is the optimal single variable selection. For the optimal pair of variable selection and because of the Bellman principal, the modeler do not need to test all twenty-one sampling distributions, but only six pairs of  variables that includes age $X_1$, in which case diastolic blood pressure is the second variable that joins the variable hierarchy. Similarly, for the optimal triple of variable, the modeler do not need to test all thirty-five sampling distributions, but only the five triples of variables that includes both age $X_1$ and diastolic blood pressure $X_4$, in which case the number of smoked cigarettes per day $X_7$ is the third variable that joins the variable hierarchy. This recursive process is repeated until there is no increase in the considered performance metric, e.g., the true positive rate. In other words, this recursive process stops when the addition of any new variable in an optimal sub-hierarchy leads to a decrease in the considered performance metric.

\subsection{Framingham CHD Analysis per Sex}
\begin{figure}[htbp]
\begin{center}
\includegraphics[scale=.3]{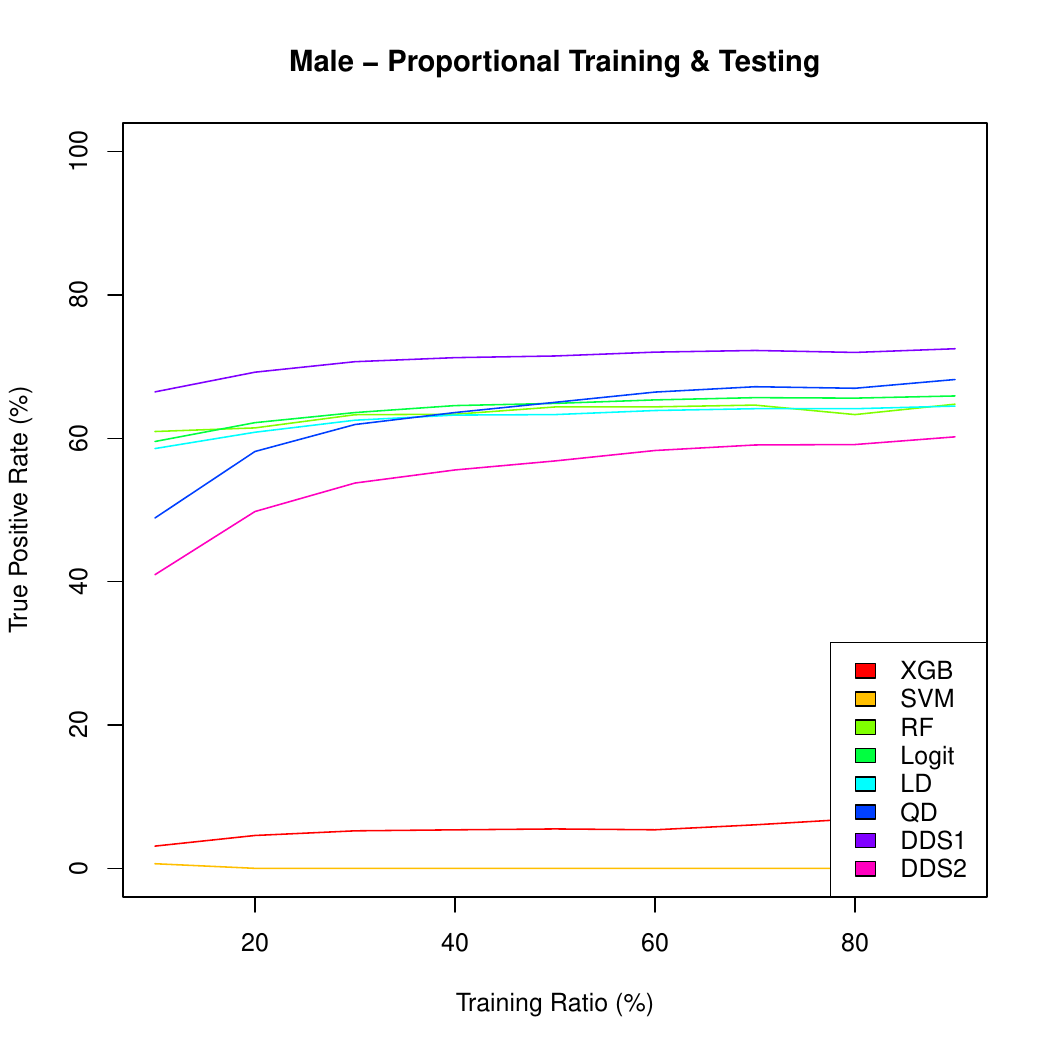}\includegraphics[scale=.3]{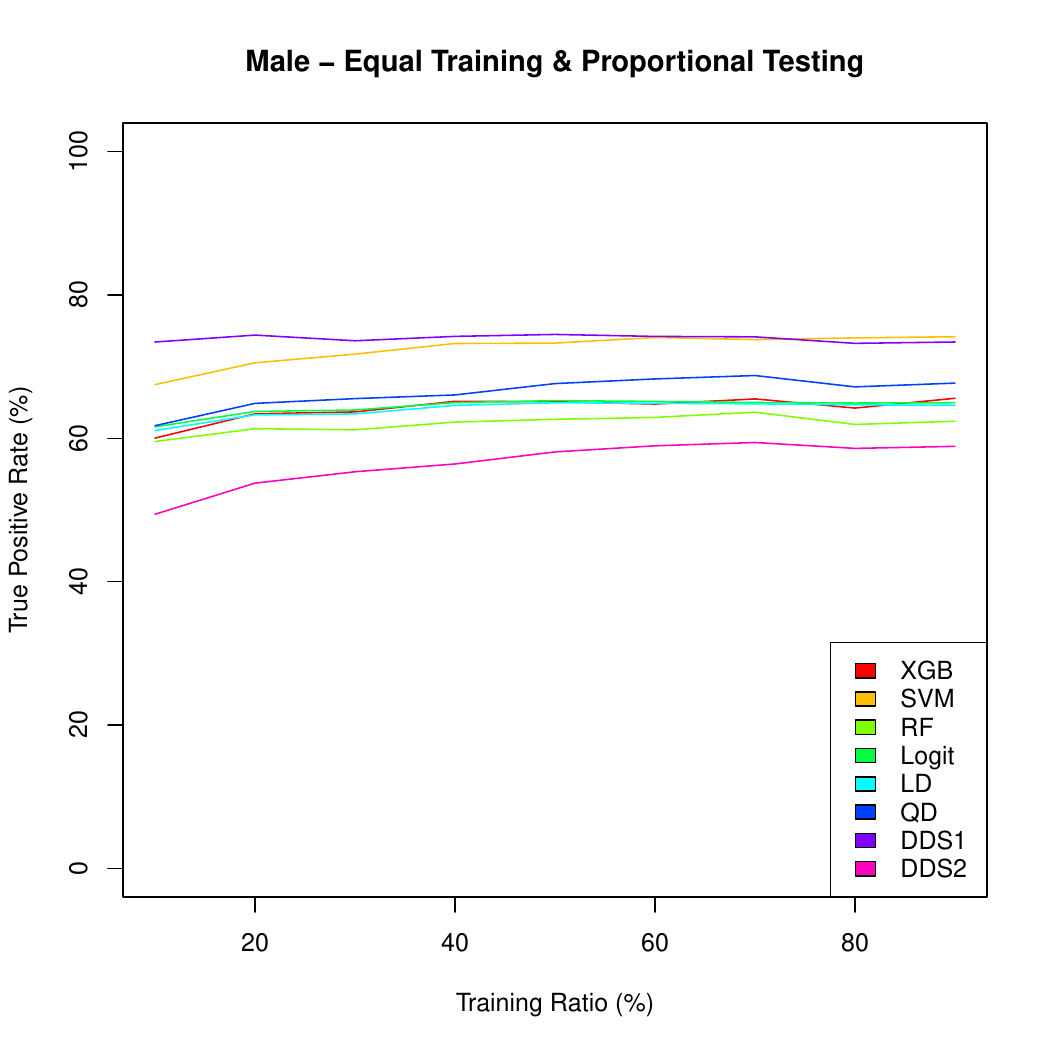} \includegraphics[scale=.3]{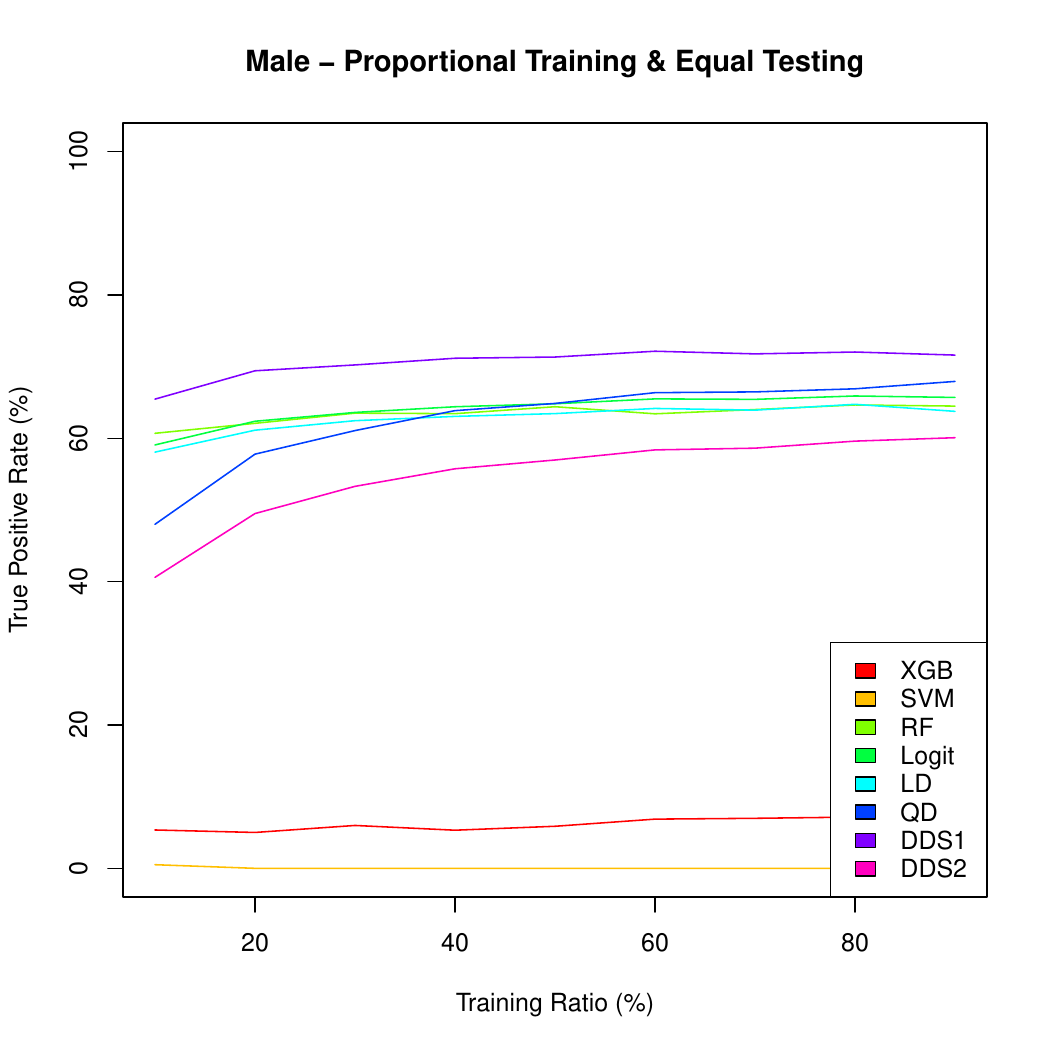}\includegraphics[scale=.3]{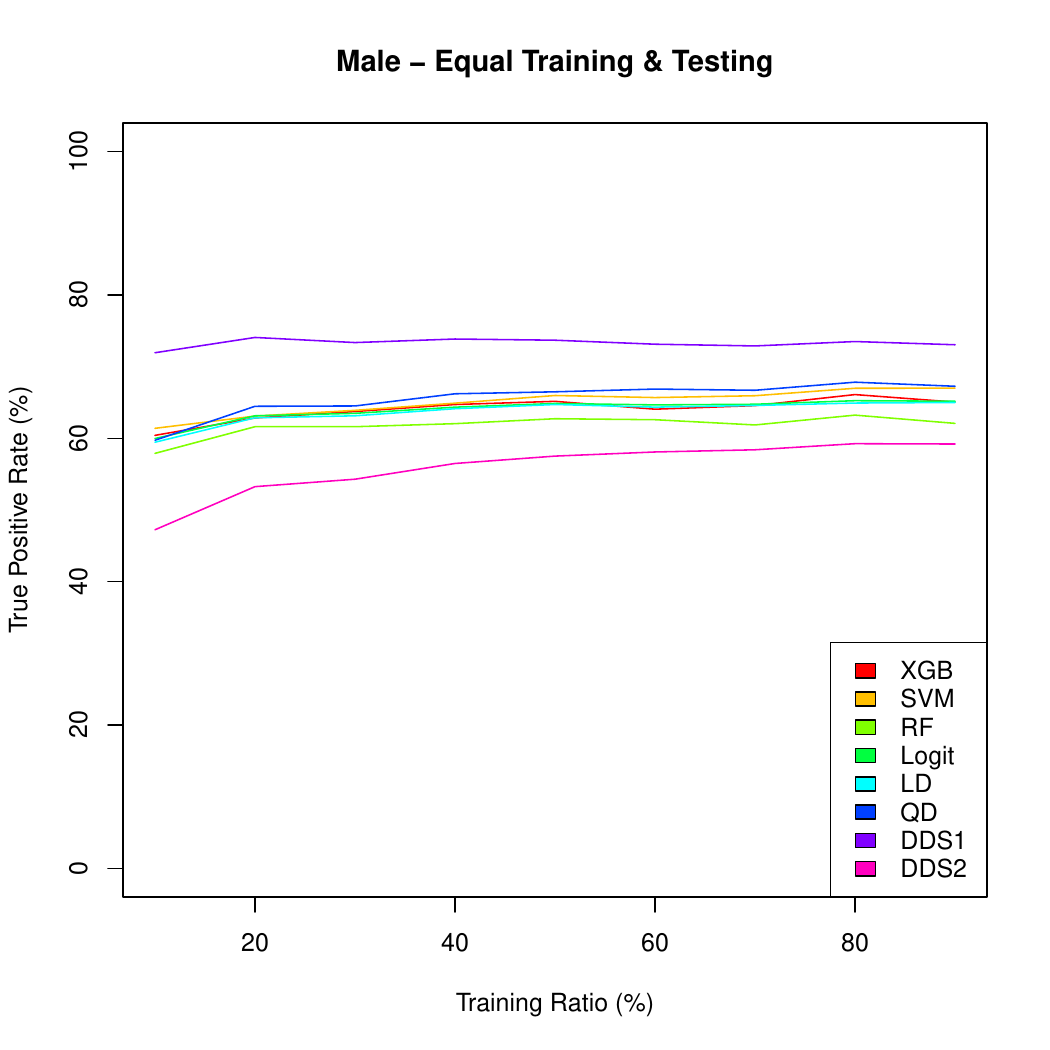}
\includegraphics[scale=.3]{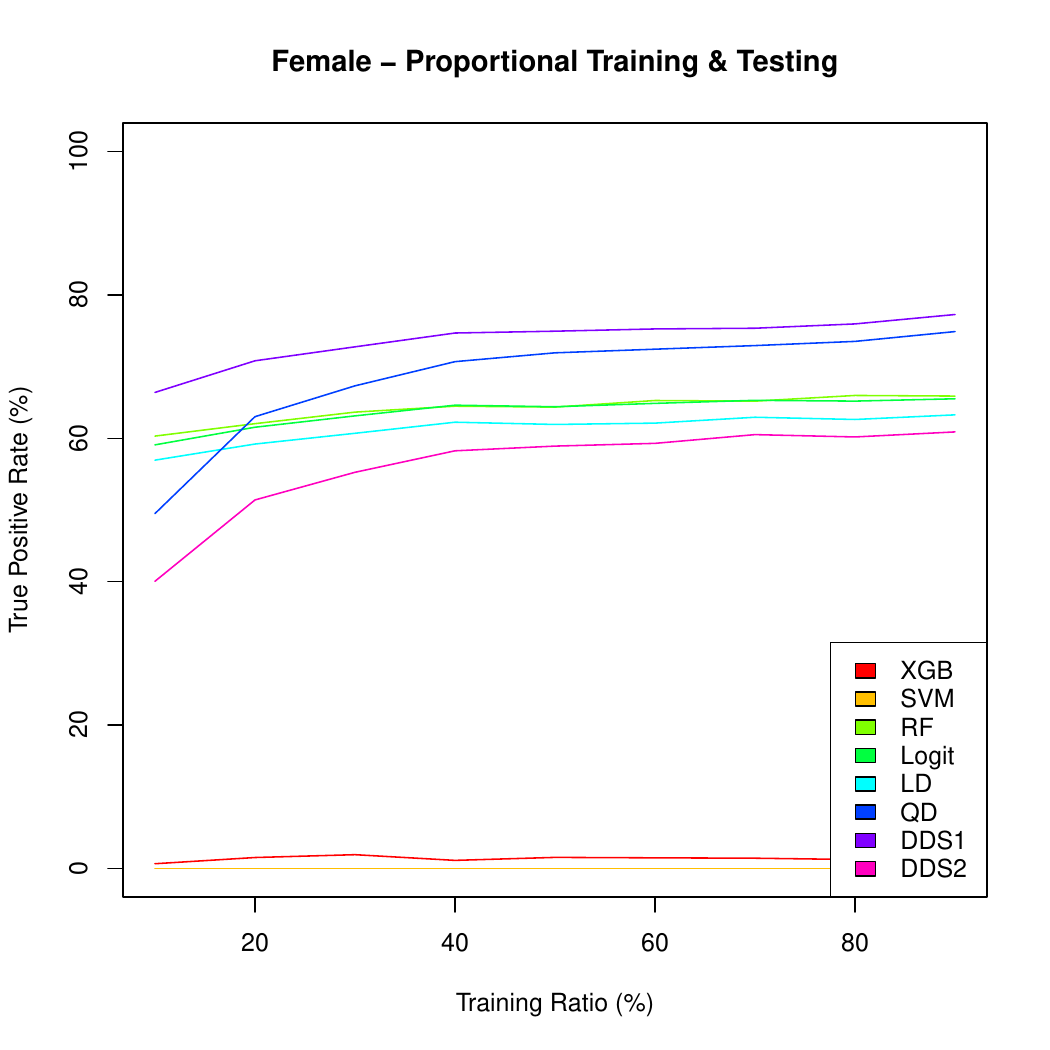} \includegraphics[scale=.3]{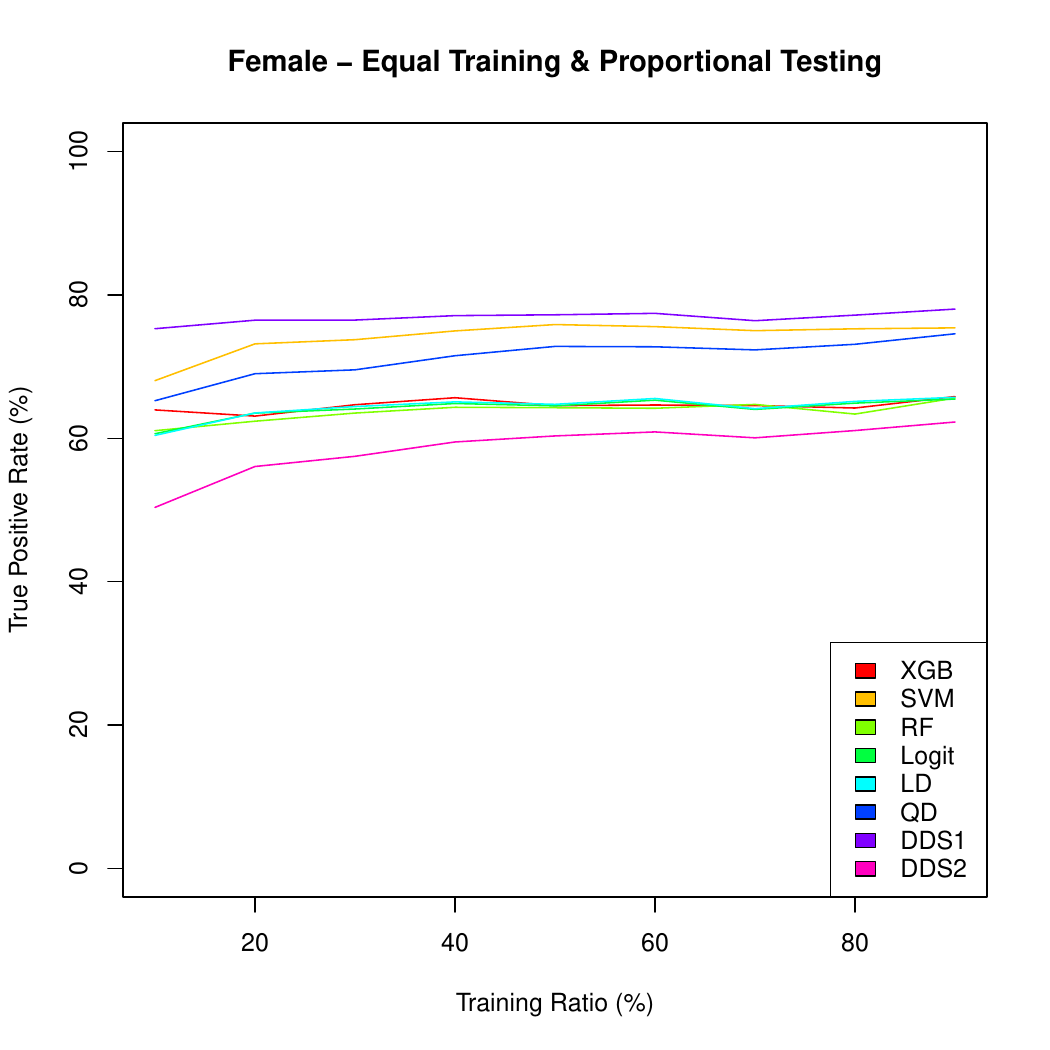} \includegraphics[scale=.3]{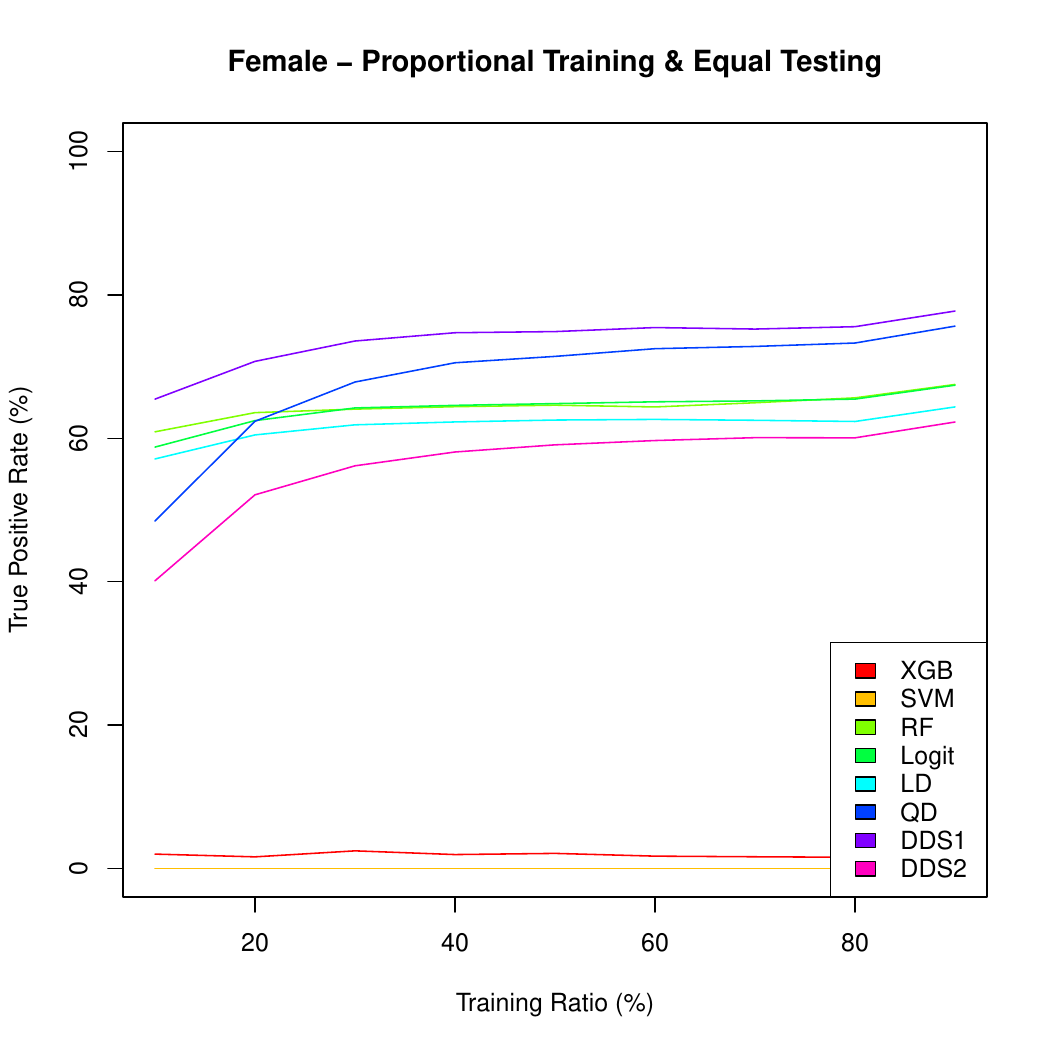}\includegraphics[scale=.3]{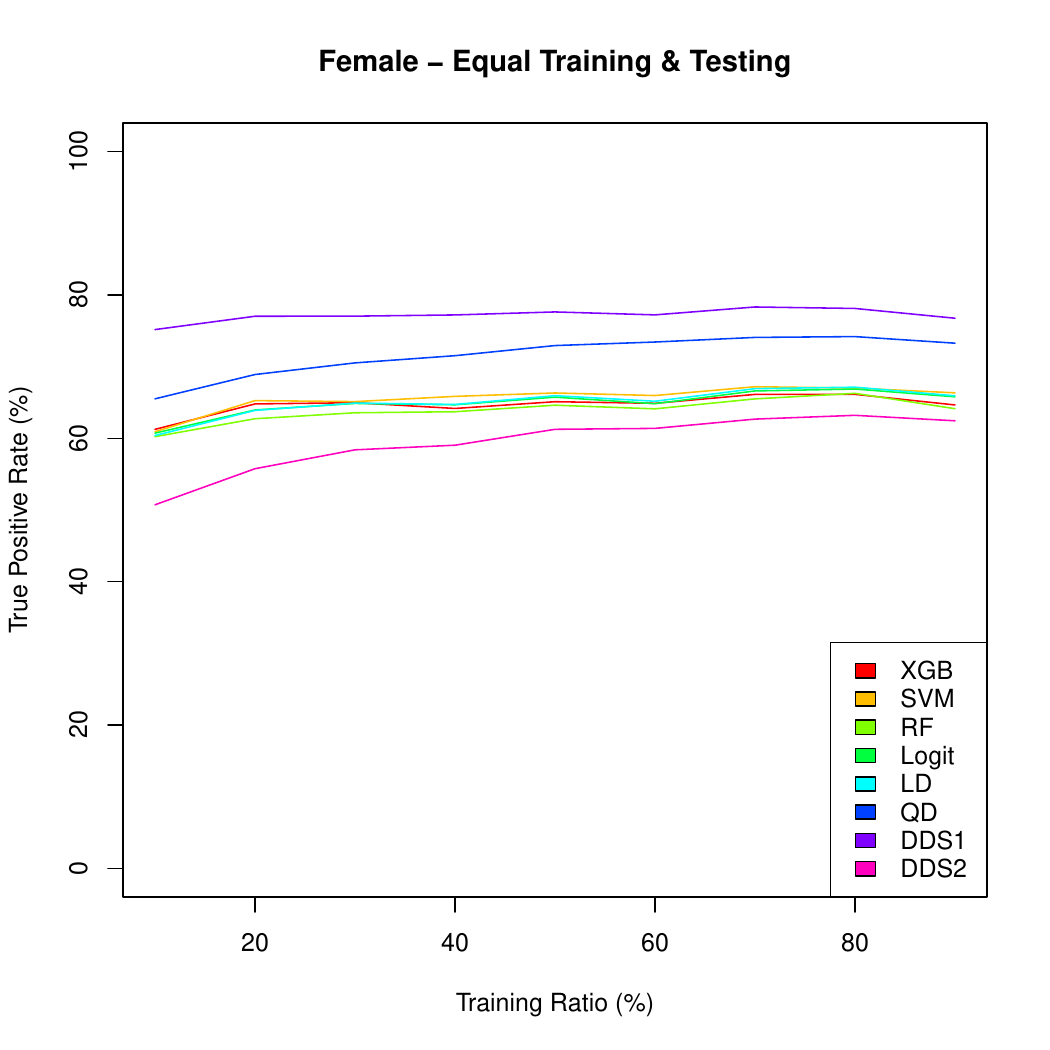}
\caption{Mean of 100 True Positive Rates for eight classification algorithms as functions of the training ratio across four training/testing scenarios for the Framingham CHD Male and Female Data}
\label{TS8algoTPRMF}
\end{center}
\end{figure}

In this subsection, we perform the above analysis, i.e., training ratio analysis (Figure \ref{TS8algoTPRMF}), classification algorithm comparison, and the double discriminant scoring methodology on the Framingham CHD data per sex (Table \ref{VarHiearchyTableMF}). Out of the 622 observations in Group 1, 337 are male and 285 are female. Out of the 3520 observations in Group 2, 1456 are male and 2064 are female. Therefore, the prevalence of CHD in males is 18.80\% and the prevalence of CHD in female is 13.81\%.

\begin{table}[htp]
\begin{center}
\begin{tabular}{cllrrr}
\hline
&Training & Testing & $n_1$ & $n_2$ & $n_3$\\ \hline
\multirow{4}{*}{Male}&Proportional & Proportional &  270&  1165& 67 + 291 = 358 \\
&Equal & Proportional & 270& 1165& 67 + 291 = 358 \\
&Proportional & Equal &270  & 1165 & 67+67 = 134 \\
&Equal & Equal & 270 & 270  & 67+67 = 134 \\
\hline
\multirow{4}{*}{Female}&Proportional & Proportional & 228 & 1651  & 57+ 413 = 470  \\
&Equal & Proportional & 228&  228& 57 + 413   = 470  \\
&Proportional & Equal &228  &  1651& 57 + 57 = 114  \\
&Equal & Equal & 228 &  228 & 57 + 57 = 114  \\
\hline
\end{tabular}
\end{center}
\caption{Training and testing datasets sizes across four training/testing scenarios  and for a training ratio $\tau = 0.8$}
\label{default}
\end{table}%

The graphs in Figure \ref{TS8algoTPRMF}. are consistent with results Figure \ref{TS8algoTPR}. in subsection \ref{ddsm}. Indeed, both the Extreme Gradient Boosting and the Support Vector Machine have an extremely low true positive rates when the training datasets are proportional, and the Double Discriminant Scoring of Type 1 is consistently the best classification algorithm for all training/testing scenarios and for each of the training ratios $\tau = 0.1, 0.2, \dots, 0.9$. In light of these results, we perform the sampling distribution analysis for the Double Discriminant Scoring of Type 1, and the optimal variable hierarchy per sex and for all four training/testing scenarios, whose results are summarized in  Table \ref{VarHiearchyTableMF}.

\begin{table}[htbp]
\begin{center}
\begin{tabular}{cllllc}
\hline
\multirow{2}{*}{Sex}&  \multirow{2}{*}{Training} & \multirow{2}{*}{Testing}   &  Optimal  & Mean \%  \\
&&& Variable Hierarchy &  True Positive Rate \\\hline
\multirow{4}{*}{Male} &  
Proportional& Proportional & (1,2,4,7,5,6)  & 73.86820\\
& Equal & Proportional & (1,2,4,7,5,6)  & 74.77845 \\
 & Proportional &Equal & (1,2,4,7,5,6) & 73.97563\\
 &Equal & Equal & (1,2,4,7,5,6)  & 74.98736 \\
\hline

\multirow{4}{*}{Female} &  
Proportional& Proportional & (1,4,2,5,7,6)  & 78.82797\\
& Equal & Proportional & (1,2,4,5,6,7)  &  79.35948\\
 & Proportional &Equal & (1,4,5,2,7,6) &78.81030 \\
 &Equal & Equal & (1,2,4,5,7,6)  & 78.41210 \\
\hline
\end{tabular}
\end{center}
\caption{Optimal variable hierarchy of the double discriminant scoring of type 1 per sex and across all four training/testing scenarios}
\label{VarHiearchyTableMF}
\end{table}%

For both male and female data, and with respect to the true positive rate, the optimal variables for predicting CHD are all variables but systolic blood pressure, which is the same as for the overall Framingham data. However, the hierarchies of these six optimal variables are different. Indeed, the optimal variable hierarchy for the predicting CHD for male patient is the sequence age $X_1$,  total cholesterol $X_2$, diastolic blood pressure $X_4$, number of cigarettes smoked per day $X_7$, BMI $X_5$ and heart rate $X_6$. This optimal variable hierarchy is the same across all training/testing  scenarios for the Framingham CHD male data with a mean true positive rates between 73.87\%  and 74.99\%. The optimal variable hierarchy for the predicting CHD for female patients, while having the same six variables, differs depending on the training/testing scenarios, as shown in Table \ref{VarHiearchyTableMF}. For all training/testing scenarios, age $X_1$ is the first variable in the hierarchy. The second variable in the hierarchy is either total cholesterol $X_2$ when the training datasets are equal, and diastolic blood pressure $X_4$ when the training datasets are proportional. The third and fourth variables in the hierarchy are between total cholesterol $X_2$, diastolic blood pressure $X_4$ and the BMI $X_5$. Finally, the fifth and sixth variables in the hierarchy for the Framingham CHD female data is either the heart rate $X_6$ or the number of cigarettes smoked per day. The mean true positive rate for the Framingham CHD female data is between 78.41\% and 79.36\%.

We conclude this subsection by giving the mean (of 1000 simulations)  classification performance matrices (Tables 
\ref{MCPMAllEqPro},  \ref{MCPMMaleEqPro}, \ref{MCPMFemaleEqPro})
  for all, male and female CHD data using the optimal variable hierarchy $(X_1, X_2, X_4, X_5, X_6, X_7)$, for equal training and proportional testing, and training ratio $\tau = 0.8$.


\begin{table}[htp]
\begin{center}
\begin{tabular}{cc|rr|cc}
\cline{3-4}
&&\multicolumn{2}{c|}{Predicted}& \\ \cline{3-6}
&& Positive &Negative & \multicolumn{1}{c|}{Total} & \multicolumn{1}{c|}{True Rate \%} \\ \hline
\multicolumn{1}{|c}{\multirow{2}{*}{Actual} }& \multicolumn{1}{|c|}{Positive} 
&97.38  & 26.62& \multicolumn{1}{c|}{124} & \multicolumn{1}{c|}{ 78.53} \\
\multicolumn{1}{|c}{}&\multicolumn{1}{|c|}{Negative} & 
 326.35& 377.65 & \multicolumn{1}{c|}{ 704} & \multicolumn{1}{c|}{53.64} \\ \hline
&\multicolumn{1}{|c|}{Total} &
 423.73&  404.27& \multicolumn{1}{c|}{828} & \multicolumn{1}{c|}{14.98} \\ \cline{2-6}
& \multicolumn{1}{|c|}{Precision \%} &  
 23.00& 93.43  & \multicolumn{1}{c|}{51.17} & \multicolumn{1}{c|}{57.37}\\ \cline{2-6}
\end{tabular}
\end{center}
\caption{Mean Classification Performance Matrix for Framingham CHD data for the optimal variable hierarchy $(X_1, X_2, X_4, X_5, X_6, X_7)$,  equal training and  proportional  testing and training ratio $\tau = 0.8$. }
\label{MCPMAllEqPro}
\end{table}%

\begin{table}[htp]
\begin{center}
\begin{tabular}{cc|rr|cc}
\cline{3-4}
&&\multicolumn{2}{c|}{Predicted}& \\ \cline{3-6}
&& Positive &Negative & \multicolumn{1}{c|}{Total} & \multicolumn{1}{c|}{True Rate \%} \\ \hline
\multicolumn{1}{|c}{\multirow{2}{*}{Actual} }& \multicolumn{1}{|c|}{Positive} &
  50.30 & 16.70& \multicolumn{1}{c|}{67} & \multicolumn{1}{c|}{75.07} \\
\multicolumn{1}{|c}{}&\multicolumn{1}{|c|}{Negative} & 
135.08& 155.92 & \multicolumn{1}{c|}{ 291} & \multicolumn{1}{c|}{53.58 } \\ \hline
&\multicolumn{1}{|c|}{Total} &
 185.38& 172.62  & \multicolumn{1}{c|}{358} & \multicolumn{1}{c|}{18.72} \\ \cline{2-6}
& \multicolumn{1}{|c|}{Precision \%} &  
 27.17& 90.36 & \multicolumn{1}{c|}{51.78 } & \multicolumn{1}{c|}{57.60}\\ \cline{2-6}
\end{tabular}
\end{center}
\caption{Mean Classification Performance Matrix for Framingham CHD Male data for the optimal variable hierarchy $(X_1, X_2, X_4, X_5, X_6, X_7)$, equal training and proportional testing and training ratio $\tau = 0.8$. }
\label{MCPMMaleEqPro}
\end{table}%

             \begin{table}[htp]
\begin{center}
\begin{tabular}{cc|rr|cc}
\cline{3-4}
&&\multicolumn{2}{c|}{Predicted}& \\ \cline{3-6}
&& Positive &Negative & \multicolumn{1}{c|}{Total} & \multicolumn{1}{c|}{True Rate \%} \\ \hline
\multicolumn{1}{|c}{\multirow{2}{*}{Actual} }& \multicolumn{1}{|c|}{Positive} &
  45.14 & 11.86& \multicolumn{1}{c|}{ 57} & \multicolumn{1}{c|}{79.19} \\
\multicolumn{1}{|c}{}&\multicolumn{1}{|c|}{Negative} & 
193.99&  219.01& \multicolumn{1}{c|}{ 413} & \multicolumn{1}{c|}{53.03 } \\ \hline
&\multicolumn{1}{|c|}{Total} &
239.13 &230.87  & \multicolumn{1}{c|}{470} & \multicolumn{1}{c|}{12.13} \\ \cline{2-6}
& \multicolumn{1}{|c|}{Precision \%} &  
 18.90& 94.88 & \multicolumn{1}{c|}{ 50.88} & \multicolumn{1}{c|}{56.20}\\ \cline{2-6}
\end{tabular}
\end{center}
\caption{Mean Classification Performance Matrix for Framingham CHD Female data for the optimal variable hierarchy $(X_1, X_2, X_4, X_5, X_6, X_7)$, equal training and testing and training ratio $\tau = 0.8$. }
\label{MCPMFemaleEqPro}
\end{table}%

\section{Discussion}Using the Framingham CHD data, and four training/testing scenarios, we showed that using a classifier cutoff equal to the prevalence of the training data is the best for converting both logistic and random forest regressions into a classification algorithm. Moreover, using statistically significant variables of a logistic regression doesn't improve the performance of the logistic classifier, and thus, one can either use these significant variables, or use all available variables.  A sampling distribution comparison of eight classification algorithms (Extreme Gradient Boosting, Support Vector Machine, Random Forest classifier, Logistic classifier, Linear Discriminant analysis, Quadratic Discriminant analysis, and Double Discriminant Scoring of Types 1 and 2)  through a paired design and across four training/testing scenarios led to the following results: 1) Both the Extreme Gradient Boosting and Support Vector Machine are flawed when the prevalence of the training data set are proportional, and thus these two classification algorithm must be derived with a training dataset with a 50\% prevalence, 2) The Logistic and Random Forest classifier derived using a (balanced equilibrium) cutoff (equal to the prevalence of the training data set) performs fairly consistently across all four training scenarios, and 3) the Double Discriminant Scoring of Type 1 consistently outperformed all other classification algorithm with respect to the true positive rate and across all training/testing scenarios, and hence one can be confident about its generalizability, i.e., using  a derived optimal double discriminant scoring of type 1 model  to make predictions on the same type of data even with different distribution, for instance, a medical data from a different geographical location. Furthermore, the sampling distribution of the performance of the Double Discriminant Scoring of Type 1 across all training/testing scenarios has the Bellman principle, and thus a modeler do not need to perform a sampling distribution test of $2^p - 1$ variables for a $p$-variable data, but would need at most $p(p+1)/2$ tests using a recursion, that is, find the optimal one variable selection, then add the other $p-1$- variables and check whether the used performance metric improved, in which case we repeat the process until the addition of a variable to the variable hierarchy no longer increases the considered performance metric. For instance, out of 127 total sampling distribution tests, 28  were needed for the optimal variable hierarchy with respect to the true positive rate, which lead six variables out of seven, and 13 tests were needed for the optimal variable hierarchy with respect to the true negative rate, which lead one variable out of seven. Moreover, both optimal variable hierarchies for the true positive and negative rates are complementary, leading us to conjecture that the missing variables in the optimal hierarchy with respect to the true positive rate belong to the optimal variable hierarchy with respect to the true negative rate.





\vspace{6pt}

\subsection*{Funding}This research is funded by the National Science Foundation (NSF DMS-2331502).

\subsection*{Data Availability}Framingham Coronary Heart Disease data is publicly available on Kaggle.   




\subsection*{Acknowledgments}Part of this research was performed while the author was visiting the Institute for Mathematical and Statistical Innovation (IMSI), which is supported by the National Science Foundation (Grant No. DMS-1929348).  Moreover, research reported in this publication was partially supported by the National Institutes of Health’s National Cancer Institute, Grant Numbers U54CA202995, U54CA202997, and U54CA203000. The content is solely the responsibility of the authors and does not necessarily represent the official views of the National Institutes of Health.


\subsection*{Conflicts of Interest}The author declares no conflict of interest.








\vspace{.35in}

\noindent

\address{Department of Mathematics\\ Northeastern Illinois University\\
Chicago, IL 60625-4699, USA\\
\email{n-kahouadji@neiu.edu}}
\end{document}